# Visual representations in the human brain are aligned with large language models

Adrien Doerig[1,2+], Tim C. Kietzmann[1+], Emily Allen[3,4], Yihan Wu[5],
Thomas Naselaris[6], Kendrick Kay[3*], & Ian Charest[7,8*]

[1]Institute of Cognitive Science, University of Osnabrück, Germany
[2]Donders Institute for Brain, Cognition & Behaviour, Nijmegen, the Netherlands
[3]Center for Magnetic Resonance Research (CMRR), Department of Radiology, University of Minnesota, Minneapolis, MN, USA.
[4]Department of Psychology, University of Minnesota, Minneapolis, MN, USA
[5]Graduate Program in Cognitive Science, University of Minnesota, Minneapolis, MN, USA.
[6]Department of Neuroscience, University of Minnesota, Minneapolis, MN, USA.
[7]cerebrUM, Département de Psychologie, Université de Montréal, Montréal, Canada
[8]Center for Human Brain Health, School of Psychology, University of Birmingham, Birmingham, United Kingdom

+ shared first authors; * shared last authors; corresponding author: ian.charest@umontreal.ca

**Abstract**

The human brain extracts complex information from visual inputs, including objects, their spatial and semantic interrelations, and their interactions with the environment. However, a quantitative approach for studying this information remains elusive. Here, we test whether the contextual information encoded in large language models (LLMs) is beneficial for modelling the complex visual information extracted by the brain from natural scenes. We show that LLM embeddings of scene captions successfully characterise brain activity evoked by viewing the natural scenes. This mapping captures selectivities of different brain areas, and is sufficiently robust that accurate scene captions can be reconstructed from brain activity. Using carefully controlled model comparisons, we then proceed to show that the accuracy with which LLM representations match brain representations derives from the ability of LLMs to integrate complex information contained in scene captions beyond that conveyed by individual words. Finally, we train deep neural network models to transform image inputs into LLM representations. Remarkably, these networks learn representations that are better aligned with brain representations than a large number of state-of-the-art alternative models, despite being trained on orders-of-magnitude less data. Overall, our results suggest that LLM embeddings of scene captions provide a representational format that accounts for complex information extracted by the brain from visual inputs.

**Introduction**

The visual system provides the brain with a wealth of information about the physical environment. Much progress in understanding the functional organisation[1–5] and computational principles[6–9] of the visual system has been driven by a heavy focus on the objects that are present in visual scenes. In particular, exciting advances in the ability to quantitatively predict neural activity in extrastriate visual cortex have been achieved by training artificial neural networks (ANNs) to perform object recognition from raw visual inputs[10–14].

Despite this progress, it is clear that visual scenes convey more information than the identity of the objects present[15]. Presumably, an effective interpretation of a visual scene must include the context in which objects reside as well as their spatial and semantic interrelations. Studies of the neural basis of object context and relations have provided insight into the role of object co-occurrence statistics[16,17], spatial and semantic interrelations among objects[18–21], the context in which objects appear[22], and their typical locations in scenes[23–26]. In addition, a robust literature on scene representations in the brain has emerged[27,28], providing insights into scene categories[27,29–33], scene grammar[26], and action affordances[34], to name a few topics. However, it remains unclear how to connect and integrate the insights obtained from these studies with the kind of quantitative and computational methods (including image-computable models) associated with the object recognition literature. A quantitative approach for studying the complex information extracted from visual scenes seems elusive: what representational format could be used to summarise and study this information?

Excitingly, recent advances in artificial intelligence (AI) provide clues into the challenge of representing scene information. First, large language models (LLMs) have made enormous strides in natural language processing[35]. LLMs learn to encode rich contextual information and statistical world knowledge through training on massive amounts of text data[36–39]. Second, AI researchers have demonstrated improvements in the ability of vision models to segment, recognize, and generate images by aligning visual representations with the information conveyed by textual image captions[40–43]. Importantly, these image captions are transformed into a powerful operable format through embedding in the latent space of LLMs[44,45]. These insights lead to an intriguing possibility: LLM embeddings of image captions might be an effective way to capture the rich information conveyed by visual scenes.

In this paper, we explore the hypothesis that LLM embeddings of image captions can be used to characterise the transformation of visual information in the human brain. By combining 7T fMRI data collected while participants viewed natural scenes with a variety of analyses and ANN modelling, we show that the visual system may converge, across various higher-level regions, towards representations that are aligned with LLM embeddings of scene captions.

## Results

To explore representational transformations across the visual system, we take advantage of the recent Natural Scenes Dataset (NSD)[46], a large-scale 7T fMRI dataset featuring brain responses to thousands of complex natural scenes taken from the Microsoft Common Objects in Context (COCO) image database[47,48]. The COCO database includes human-supplied captions describing each image, as well as labels for object categories present in each image. To test whether LLM embeddings provide a useful representational format for modelling visually evoked brain responses, we use LLM sentence encoders based on transformer architectures[49] and project the scene captions into the embedding space of these LLMs (Fig. 1A). As a representative LLM, we use MPNet[39], a transformer that is finetuned for sentence-length embeddings. MPNet was chosen as it reaches state-of-the-art performance on a variety of benchmarks, including Semantic Textual Similarity (STS), which measures the match with human judgments of semantic similarity between sentences[50]. Importantly, our LLM embeddings are derived fully from text, without regard for visual features of the corresponding scenes. This differs from other embeddings that are jointly trained on visual input and language (e.g. Contrastive Language–Image Pre-training (CLIP)[43]). A 2D T-SNE projection of MPNet-embedded NSD captions confirms that the model successfully captures fine-grained scene information, such as what objects are present, what actions are being performed, and the type of scene (Supp. Fig. 1).

*A linear mapping from LLM embeddings captures brain responses to natural scenes*

To quantify how well LLM embeddings of scene captions predict brain activities, we used Representational Similarity Analysis (RSA)[4,51–53]. We correlated Representational Dissimilarity Matrices (RDMs) constructed from LLM embeddings of the image captions with RDMs constructed from brain activity patterns obtained while participants viewed the corresponding natural scenes (Fig. 1A). Applying RSA in a searchlight fashion, we find that the LLM embeddings are able to predict visually evoked brain responses across the entire visual system, but especially in higher level visual areas in the ventral, lateral and parietal streams (Fig. 1B; see Supp. Fig. 2 for individual subjects; see Supp. Fig. 5 for a reproduction of this result using different LLMs).

We then probed the mapping between LLM representations and brain representations using linear encoding models. We first trained an encoding model to predict individual voxel activities from LLM embeddings using cross-validated fractional ridge regression[54]. In line with the RSA results, we find that the encoding model successfully predicts variance across large parts of the visual system (Fig. 1C&D, see Supp. Fig. 3 for individual subjects). This suggests that the LLM representations of associated captions accurately capture important features of visual processing. To elaborate this point, we tested if the model can reproduce well-established tuning properties observed in cognitive neuroscience. We contrasted the predictions derived from different sentences highlighting people versus scenes (e.g. "Man with a beard smiling at the camera" vs. "A view of a beautiful landscape"). Such a contrast revealed classical tuning properties associated with people- and place-selective areas (including FFA, OFA & EBA vs. PPA & OPA; Fig. 1E; also see Supp. Fig. 4). The success of the encoding model indicates that LLM representations, despite being derived purely from text, can make accurate predictions of region-specific tuning properties of the visual cortex.

The success of the LLM representations in characterising brain activity suggests that it may be possible to accurately infer a textual description of what participants saw from brain activity alone using simple linear methods. To test for this, we trained a linear decoding model to predict LLM embeddings from fMRI voxel activities (Fig. 1F). Then, to reconstruct scene captions, we used a dictionary lookup approach[55] on a large corpus of 3.1 million captions (taken from Google Conceptual Captions[56]). As shown in Fig. 1F, we obtain remarkably accurate textual descriptions of the stimuli viewed by the participants. This highlights the appropriateness of LLM embeddings as a representational format for brain signals evoked by visual stimuli.

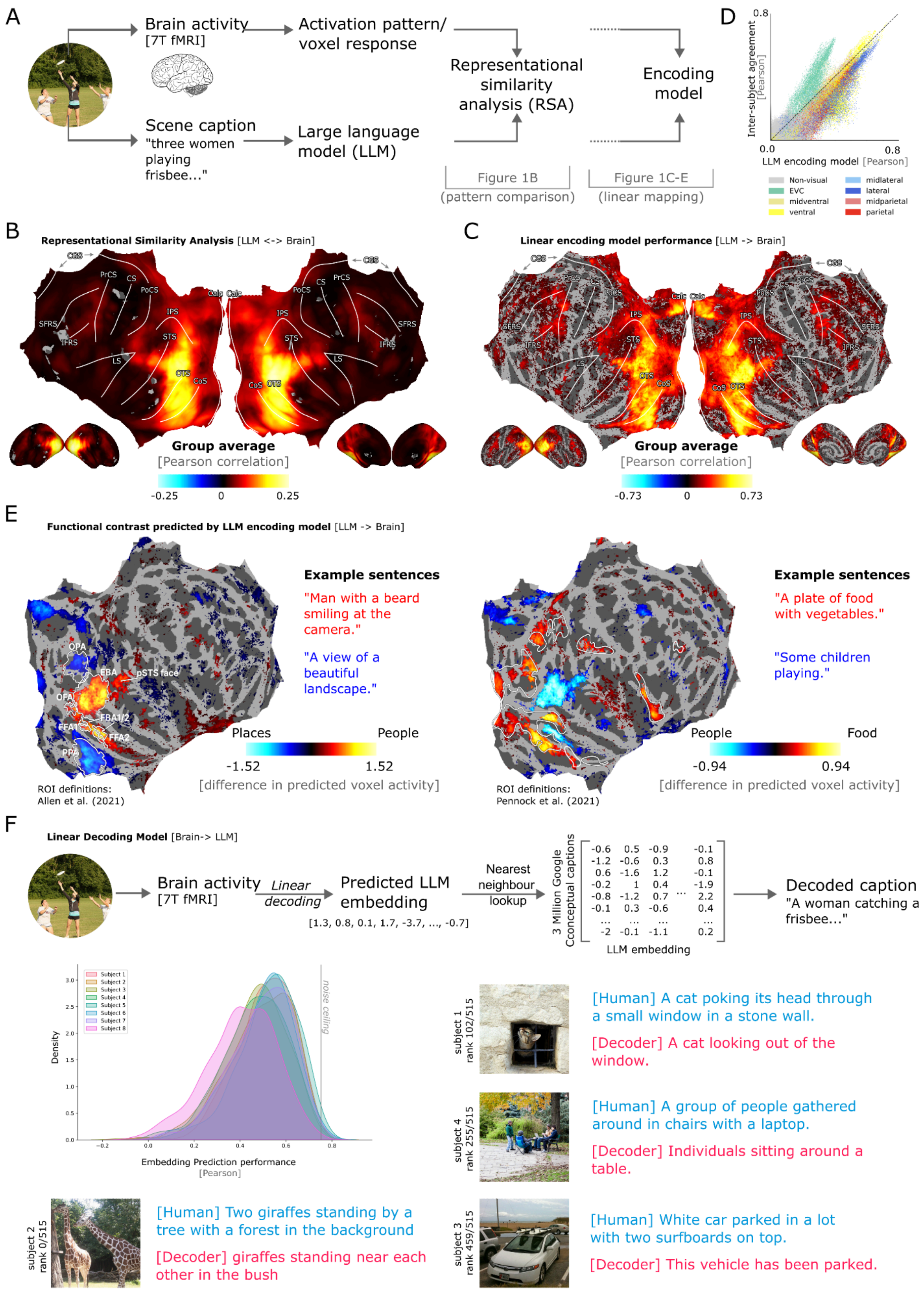

**Figure 1: A linear mapping from LLM embeddings captures visual responses to natural scenes.** ***A. LLM to brain mapping methods.*** Each image in the NSD dataset is associated with captions written by different human observers to describe the scene. These captions are passed through an LLM model to generate embeddings.

We use two approaches to quantify the match between these embeddings and fMRI data (RSA and encoding models). ***B. RSA reveals an extended network of brain regions where LLM representations correlate with brain activities.*** Searchlight map for the correlation between LLM embeddings (as given by MPNet embeddings) and brain representations. Group average (significance threshold set by a 2-tailed $t$-test across subjects with Benjamini & Hochberg false discovery rate correction; $p = 0.05$). See Supp. Fig. 2 for individual subjects. ***C. A linear encoding model highlights a similar network of brain regions.*** We performed voxel-wise linear regression to predict voxel activities from LLM embeddings. Shown is the Pearson correlation map between the predicted and actual beta responses on the test set (significance threshold set by a 2-tailed $t$-test across subjects with Benjamini & Hochberg false discovery rate correction; $p = 0.05$). See Supp. Fig. 3 for individual subjects. ***D. Encoding model performance vs. inter-subject agreement.*** Each dot in the scatter plot shows the encoding model performance for a given voxel vs. the inter-subject agreement, computed as the mean Pearson correlation between a subject's voxel activities and the average of the voxel activities of the remaining 7 subjects on the test images. Our encoding model approaches the inter-subject agreement in all ROIs, indicating good performance. Values above the diagonal can be explained by the fact that the model captures subject-specific variance not captured by the mean of other subjects. ***E. The linear encoding model captures selectivities of different brain regions.*** We contrasted the brain activity predicted from five people- vs. place-related sentences (left) and five food- vs. people-related sentences (right; significance threshold set by a 2-tailed $t$-test across subjects with $p=0.05$; without correction for false discovery rate). These contrasts highlight brain areas known to be selective for people, places and food (people and place areas are localised as part of NSD (left); food areas described by [57] shown as white outlines (right)). ***F. Decoding captions from visually evoked brain responses.*** *Top:* We fit a linear model to predict LLM embeddings (MPNet) from fMRI voxel activities. We then use a nearest neighbour lookup to generate a caption for each image. *Bottom left:* Kernel density estimate plot of the prediction score for each participant on a held out test set, quantified using Pearson correlation between predicted and target embedding. The noise ceiling is computed as the consistency between the 5 human-generated captions for each image. *Bottom right:* Target and decoded caption examples from different subjects on the held out test set, spanning the range of prediction scores. The rank refers to the prediction score of the shown sample (i.e., rank 1 is the best prediction for this subject, while rank 515 is the worst).

### *LLMs integrate complex information contained in scene captions that is important to match brain activities*

LLMs are capable of encoding and integrating complex contextual information. We hypothesised that this ability can, in part, explain the match of LLM embeddings to brain activities. We contrasted models that differ in their ability to encode contextual information in scene captions. We focused our analyses on regions of interest (ROIs) across the visual system, including early visual cortex (EVC) and the ventral, parietal, and lateral visual streams (using the NSD 'streams' ROI definitions). We use parameter-free RSA to estimate the representational agreement, and report $t$-test statistics after Benjamini & Hochberg false discovery rate correction with a significance threshold of $p<0.05$.

First, we tested if the ability of LLMs to align with visual cortex representations relies on more than just object category information (Fig. 2A). We compared LLM embeddings of full captions to several models that encode object category information in different ways, including binary multi-hot vectors (as provided by the COCO dataset), contextually enriched single word embeddings (including fasttext[58,59], which is based on the context of words, and GloVe[60], which is based on word co-occurence statistics), and finally LLM embeddings of concatenated category words. The latter showed significantly better alignment with brain representations than multi-hot vectors (except in the lateral ROI) and word embeddings (except fasttext in EVC). This shows that the LLM representational format allows for better predictions of brain activities, even when limited to category information. However, the LLM embeddings of full captions better predicted brain activities in all ROIs by far, indicating that part of the success of LLM mapping to visual brain data is due to its ability to integrate caption information that goes beyond categories.

Second, to further understand which aspects of the LLM embeddings drive their agreement with the brain data, we compared LLM embeddings extracted from the full image caption with embeddings obtained from a concatenation of all caption nouns or all caption verbs (Fig. 2B). In agreement with our previous findings, we find that the full caption embeddings significantly outperform the noun and verb-based embeddings across all ROIs tested, except for noun-based embeddings in EVC. Note that this comparison is a stronger test than the previous analysis of category words, as caption nouns include additional content such as scene locations.

Third, we tested if captions embeddings provide additional explanatory power beyond that of their constituent words (Fig. 2C). To this end, we compared the LLM caption embeddings with LLM, fasttext and GloVe embeddings averaged across all individual words. Again, in all ROIs, the embeddings of whole captions aligned significantly better with brain data than averaged embeddings of the individual caption words. This indicates that the contextual relations among the caption words are an important factor for the LLMs' alignment with visual representations in the brain.

Finally, to ensure that our results are not reliant on the exact LLM used for embedding the captions, we tested several other LLMs from the Sentence-Transformers leaderboard (https://www.sbert.net/index.html) and found that they all perform similarly to MPNet used here (Supp. Fig. 5; none of the statistical comparisons among LLM models were found to be significant). This finding speaks to the generality of our findings, and aligns with previous work indicating that scale can matter more than architectural differences in LLMs[61,62].

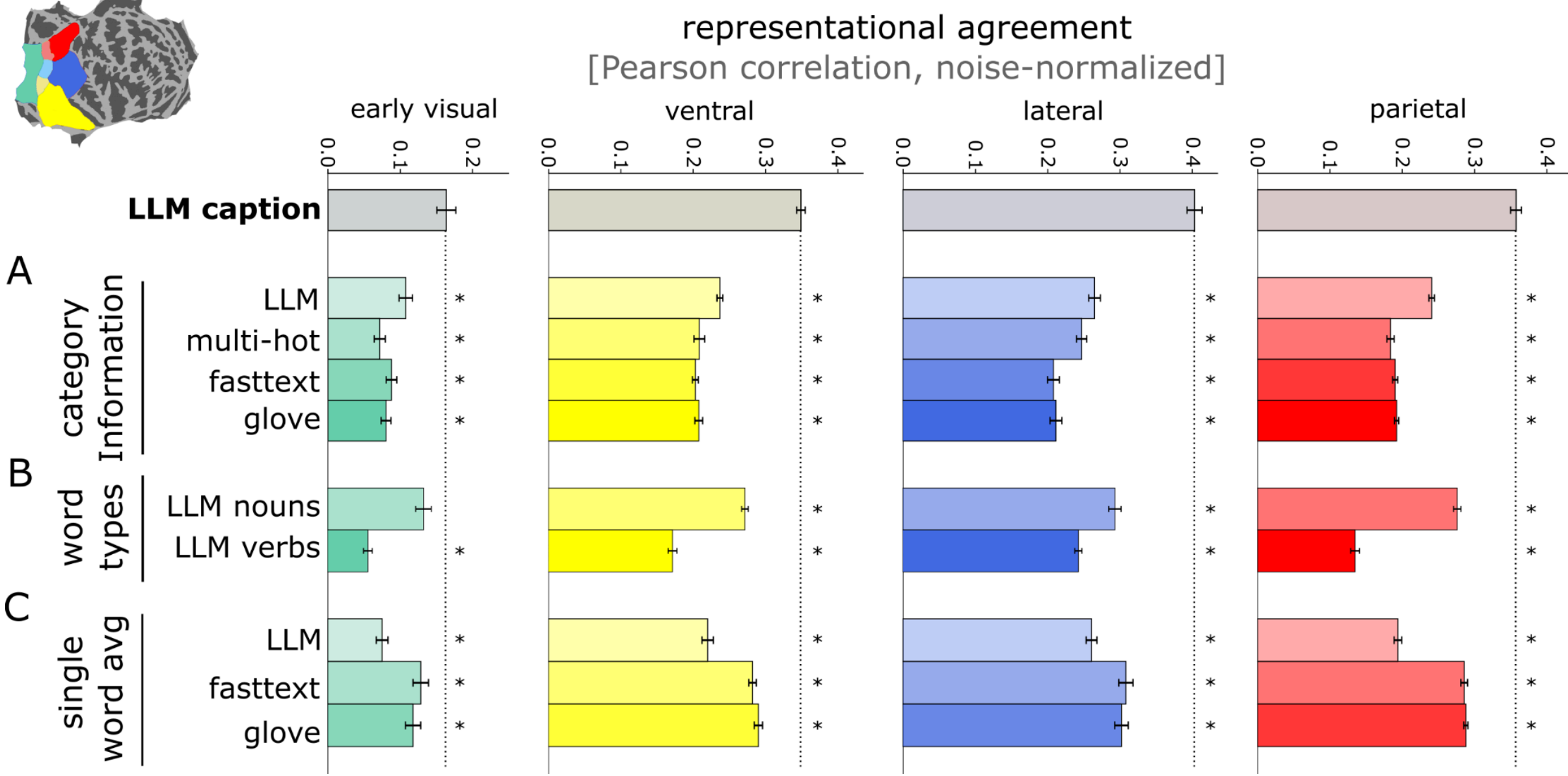

**Figure 2: The match of LLMs to visually-evoked brain activities derives from their ability to integrate complex information contained in scene captions.** We applied our RSA approach in the 'streams' ROI definitions of the NSD dataset, shown in the top-left inset. 'LLM caption' refers to the LLM embedding (MPNet) of the entire

caption, and different groups denote different classes of control models, detailed below. The match between each model and brain activities is quantified as the noise ceiling corrected correlations between each model and a given ROI (error bars reflect standard error across subjects; all statistics are 2-tailed *t*-tests across participants, with Benjamin & Hochstein correction for false discovery rate with $p$<0.05; stars show comparisons where 'LLM caption' significantly outperforms the control model; corrected *p*-values for all pairwise model comparisons are provided separately in Supp. Fig. 6). **A. LLM embeddings of category information improve match to brain data.** We compare multiple formats to represent category information, from binary multi-hot vectors (*multi-hot*), through averaging fasttext (*fasttext*) or GloVe (*glove*) word embeddings of category words, to embedding a concatenation of all category words using MPNet (*LLM*). **B. LLM embeddings capture brain-relevant information beyond nouns or verbs.** The LLM embeddings of the concatenated caption nouns (*LLM nouns*) or verbs (*LLM verbs*) both match brain data significantly less well (except *LLM nouns* in EVC) than the LLM embeddings of the full caption (*LLM caption*). **C. LLM embeddings capture brain-relevant contextual information.** To test if contextual information conveyed by captions is important to match brain data, we compared embeddings of whole captions with the averaged LLM, fasttext, and GloVe embeddings of individual caption words.

*LLM-trained RCNNs outperform other models of visual responses*

Our results indicating that brain representations are well characterised by LLM-like representations lead to the intriguing hypothesis that LLM embeddings might serve as a powerful target for training visual models. There has been a history of success using task-optimised neural network models as potential models of the visual system but, conventionally, these models are trained to classify objects present in each image[12,13,63,64] or, in some cases, using unsupervised objectives[65,66]. We therefore trained ANNs to predict LLM embeddings from visual inputs and quantified the match of these task-optimised models to our brain data (Fig. 3A).

We used recurrent convolutional neural networks (RCNNs [67]), based on the vNet architecture[63] that mirrors the progressive increase of visual field sizes across the ventral stream. The RCNNs were trained to predict LLM embeddings of the captions associated with the input scenes (LLM-trained RCNNs) on the COCO dataset. To avoid overfitting to the images seen by NSD participants, we excluded from training all images that were used in NSD. We trained 10 network instances with different random seeds to account for possible variation due to network initialisation[68]. To compare the model response to brain data, we extracted activity patterns in response to the NSD stimuli from the last layer and timestep, computed representational dissimilarity matrices, and used searchlight RSA to quantify representational alignment. This searchlight analysis revealed that the LLM-trained RCNN layer activations were able to significantly predict visually evoked brain responses across the entire visual system, similarly to the searchlight performed on the LLM embeddings themselves (Supp. Fig. 7). In fact, we found the LLM-trained RCNNs align significantly better with the brain data than the LLM embeddings they were trained to predict (Fig. 3C; Supp. Fig. 8). We interpret this finding as reflecting the fact that brain representations are not visually agnostic in these regions and retain some visual information that cannot be captured by the LLM embeddings alone (e.g. positions of objects that are not explicitly mentioned in the scene captions but are available to IT[69]). This information is captured by the LLM-trained RCNN models, which receive visual inputs.

Despite the strong correlations observed between our LLM-trained RCNNs and the brain data, it is still possible that conventional models trained to classify objects could outperform it. We therefore ran a highly controlled model comparison contrasting our LLM-trained RCNNs with RCNNs trained to predict multi-hot category labels (category-trained RCNNs; again, we trained 10 instances with different random seeds). Training these networks end-to-end enabled us to perform a stringent test of our hypothesis: both LLM-trained and category-trained RCNNs are fed the exact same images, have the exact same architecture, the same dimensionality, and the same random seeds. They differ only in their training objective (Fig. 3A). To adjudicate between the two models, we contrasted their representational alignment using RSA (focussing on the last layer and timestep activities, as previous work has shown that these layers perform best in predicting higher-level visual regions[63]; searchlight maps for all other layers and timesteps can be found in Supp. Fig. 10). In line with our hypothesis, the LLM-trained RCNNs significantly outperformed the category-trained controls across a wide network of higher visual areas (Fig. 3D). The same result was replicated using a ResNet50[70] architecture, showing that the benefit of LLM-training is not restricted to our particular RCNN architecture (Supp. Fig. 11).

These findings are still consistent with the discovery of object categories as a major factor in ventral stream representations[4,8,71–74]. Indeed, because LLM embeddings capture many forms of linguistically expressible content, LLM representations may encompass content conveyed by object category information. To assess this hypothesis, we froze the weights of our LLM-trained and category-trained RCNNs, and quantified how well category labels and LLM embeddings could be linearly read out (Fig. 3B). We found that category labels could successfully be read out from LLM-trained RCNNs (i.e. similar performance as for the category-trained RCNNs). However, the reverse was not true: LLM embeddings could not be read out from category-trained RCNNs as well as from LLM-trained RCNNs. These results suggest that the LLM representational format encompasses categorical information while providing a richer training target that improves the match to visually evoked brain activity.

We assessed our LLM-trained RCNNs in the broader landscape of ANN modelling by comparing against 13 models previously reported to be good predictors of visual activity in the brain. These models have diverse architectures, training datasets, and objectives, and include leading models on neural data prediction benchmarks such as NSD[75] and brainscore[76], supervised category-trained models (including a larger version of our RCNN architecture trained on ecoset[63], and several different models trained on Imagenet 1000k[77]), supervised models trained for scene categorization on the Places365[78] and taskonomy[79] datasets, weakly supervised models trained on hundreds of millions of images[80] or image-text pairs (CLIP[43]), and unsupervised models trained using simCLR[81] and instance-level contrastive learning[65] (see Methods for the full list of models). Notably, all of these models are trained on >1M images (ecoset/ImageNet), or hundreds of millions of images (in the case of resnext101_32x8d_wsl and CLIP), while our LLM-trained RCNN is trained on orders of magnitude less data (the 48k images left in COCO after removing NSD images).

We applied the same RSA approach as before, and report the correlation between each model's pre-readout RDMs and brain RDMs obtained from higher-level ROIs of the ventral, lateral and parietal visual streams (except for CLIP, for which we used the final embedding instead of the pre-readout layer). We find that our LLM-trained RCNN models, trained to map from pixels to LLM embeddings, significantly outperform every single other model in the ventral and parietal ROIs, and all but one (which is worse, but not significantly) in the lateral ROI (Fig. 3E). To rule out the possibility that this good alignment to brain representations is driven by training on our subset of COCO rather than by the LLM objective, we verified that RCNNs trained to predict category labels on ecoset are not outperformed by our RCNNs trained to predict category labels on our subset of COCO (and both are outperformed by our LLM-trained RCNN; again, we reproduced this result using a ResNet50 architecture; Supp. Fig. 12). This suggests that enormous training datasets may not be needed for good models of visual brain responses if a powerful objective is used, and that LLM-training is such a powerful objective to train brain-aligned ANNs. In addition, our LLM-trained RCNNs provide a proof of concept for how the brain may compute LLM-aligned representations from visual inputs.

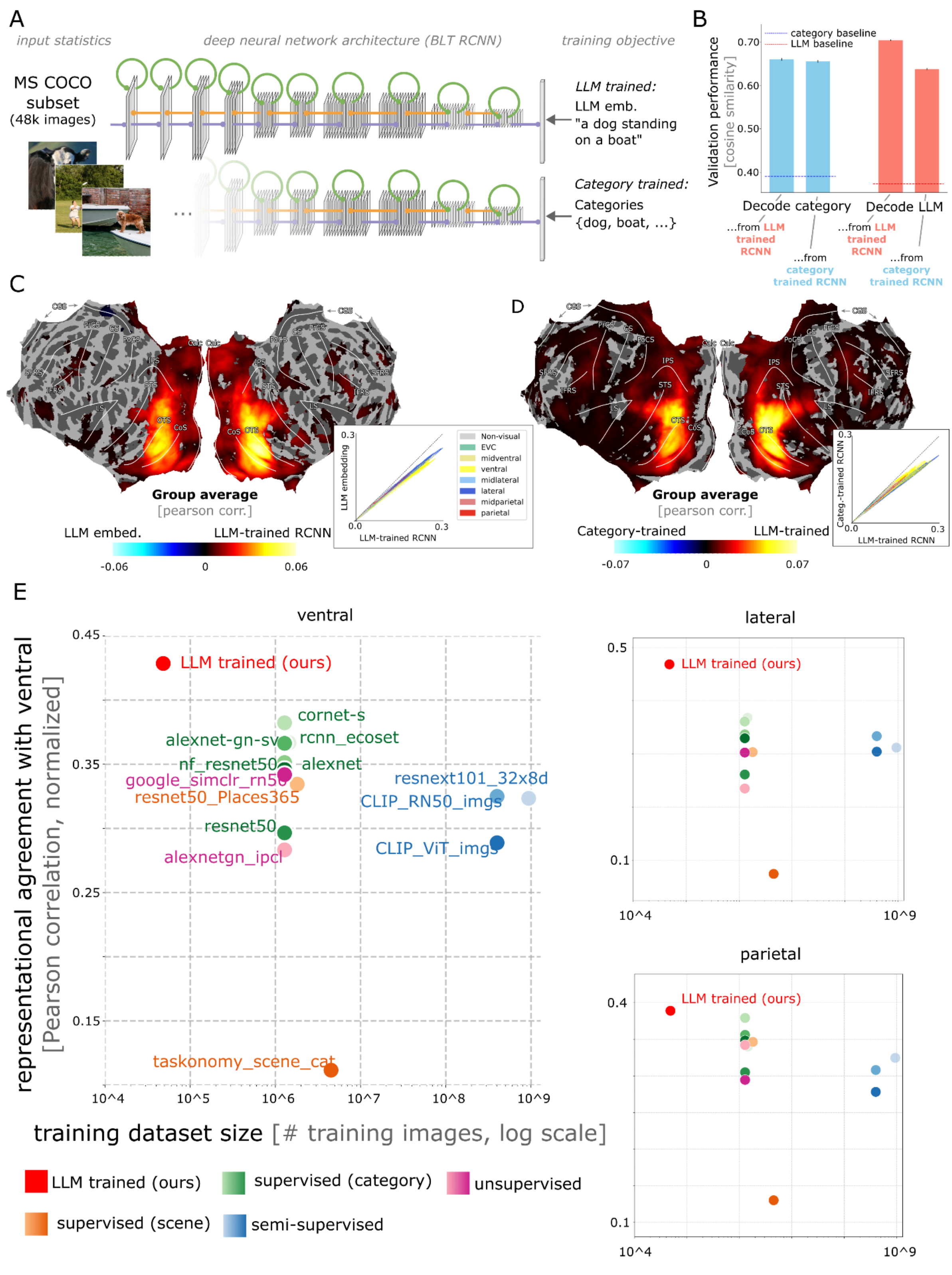

**Figure 3: LLM-trained Deep Recurrent Convolutional Networks outperform other models in predicting brain activity.** ***A. RCNN architecture*.** Our RCNN model is based on the vNet architecture[82], with ten fully recurrent convolutional layers with bottom-up (purple), lateral (green) and top-down (orange) connections, followed by a

fully connected readout layer. The receptive field sizes are chosen to be as close as possible to human foveal receptive field sizes. The training objective is to minimise the cosine distance between the network's output and the target LLM caption embeddings. Category-trained control networks are identical in all respects, except that they are trained to predict multi-hot category labels. ***B. Category labels can be read out of LLM-trained RCNN activities.*** After freezing network weights, we used a finetuning approach to test how well category labels (resp. LLM embeddings) can be decoded from activities in the pre-readout layer of the LLM-trained (resp. category-trained) network. The plot shows performance on a held out set of images, quantified as the cosine similarity between the predicted and target vector. The dashed horizontal bars show floor performance, operationalized as the performance obtained by predicting the mean training target. Category labels can be decoded from the same accuracy for the LLM-trained and category-trained networks, showing that LLM-training is rich enough to subsume category-training. In contrast LLM embeddings cannot be decoded well from the category-trained network. ***C. Searchlight RDM maps for LLM-trained RCNN vs. LLM embeddings*.** Searchlight RSA contrast between the RCNN activities in the last layer and timestep and the LLM embeddings of scene captions. The LLM-trained RCNNs significantly outperform the embeddings it was trained to predict in most of visual cortex. The insert shows a scatter plot, for each searchlight location, of the brain-model correlation for LLM–trained RCNNs vs. LLM embeddings. RCNN RDMs are averaged across 10 network instances; correlations are averaged across 8 participants; significance threshold set by a 2-tailed *t*-test across subjects with Benjamini & Hochberg false discovery rate correction; $p$ = 0.05. See Supp. Fig. 8 for individual subjects. ***D. Searchlight RDM maps for LLM-trained vs. category-trained RCNNs*.** Similar plot as panel C, but showing the contrast between LLM-trained and category-trained RCNNs (last layer & timestep). The LLM-trained RCNNs significantly outperform the category-trained controls across large parts of higher-level visual cortex. See Supp. Fig. 9 for individual subjects; Supp. Fig. 10 for all other RCNN layers and timesteps; Supp. Fig. 11 for a reproduction of this effect using the ResNet50 architecture. ***E. ROI-wise comparison of LLM-trained RCNNs with other widely used ANNs*.** Scatter plot of representational agreement between different models and the ventral, lateral and parietal ROIs, showing noise ceiling corrected correlations between the pre-readout layer and ROI RDMs. Our RCNN model significantly outperforms all other models (except CORnet-S, which is not significantly worse in the parietal ROI; 2-tailed *t*-test across participants with Benjamin & Hochstein false discovery rate correction; $p$=0.05) despite being trained on orders of magnitude fewer examples. Benjamin & Hochstein false discovery rate corrected p-values for all pairwise model comparisons are given in Supp. Fig. 13.

## Discussion

Using a variety of techniques, including RSA, encoding models, linear decoding, and ANN modelling, we have provided evidence for the hypothesis that the visual system may converge, across various higher-level regions, towards representations that are aligned with LLM embeddings of captions describing visual inputs. Furthermore, we have shown that the success of LLMs in matching brain activities comes from their ability to integrate complex information conveyed by entire captions. The robust and structured mapping between LLM embeddings and visually evoked activities paves the way for new approaches seeking to characterise complex visual information processing in the brain.

Our results build on, and extend, previous research showing the extent of features extracted by visual processing, including object[6–9] and scene[27,29–33] categories, aspects of linguistics[83,84], object occurrence statistics[17,33], the typical location of objects in scenes[23,25], and many others[24,26,27,32,85–95]. Our approach based on LLM embeddings should not be seen as a competitor to these lines of work, but rather as synergistic. For example, our results indicate that LLM embeddings subsume category information. Future work is needed to assess which other known aspects of visual processing are well captured by LLM embeddings.

The success of LLM caption embeddings in predicting visual responses to natural scenes does not imply that these embeddings fully account for the information present in brain responses. Adding access to the actual images seen by the participants can improve prediction performance, as shown by our finding that LLM-trained ANNs taking visual inputs are better aligned with the brain than the LLM caption embeddings. Our interpretation is that the visual system encodes visual input into a representational format that aligns with LLM caption embeddings while retaining some visual information. This interpretation is supported by the good performance of our ANNs, that predict LLM embeddings from visual inputs, compared to a wealth of control models (see [10,11,96,97] for discussions of this approach of contrasting ANN models to test computational hypotheses about brain processing).

The fact that our LLM-trained ANNs outperform a wealth of state-of-the-art neuro-AI models in predicting visually-evoked brain activity may seem surprising, as they are trained from scratch on a dataset that is orders of magnitude smaller. Large, varied datasets have been cited as one of the most important ingredients to build ANN models that perform well in predicting neural activities and behaviour in the visual system[75,98]. Our results suggest that huge visual datasets may not be required, if a powerful training objective is used. By training on large-scale linguistic corpora (which is thought to be crucial for learning good embeddings[61,62]), LLMs encode a wealth of contextualised, and behaviourally-relevant information. This rich learning signal derived from language might provide important benefits over other training objectives, including the supervised, unsupervised, and weakly-unsupervised approaches we tested.

Our results do not imply that visual representations have all distinctive attributes of language, such as recursivity and syntax. Rather, what we show is that LLM representations of pure textual input show strong alignment with higher level visual representations. This is in line with recent work in AI, which showed that LLMs can be used to improve the representations of visual models[40–45], as well as neuroscientific work highlighting similarities between linguistic and visual representations in the brain[99] and showing that linguistic information improves the ability of crossmodal ANNs to predict brain activities[100,101].

The NSD participants were tasked to report if they had previously seen each presented image. It cannot be fully ruled out that, to perform this continuous recognition task, participants were internally captioning the scenes, and this may have benefited the LLM caption embeddings as a good model of visual responses. Alternatively, the brain responses may align well with LLM caption embeddings irrespective of task demands. While data of the scale of NSD is currently not available for other task settings[102], it will be interesting for future work to investigate LLM-based codes under different tasks. For example, one could use encoding models, as done here, to map from high-level LLM embeddings to visual responses obtained while participants engage in different tasks, and investigate the loadings of the linear model on different embedding dimensions[103].

A representational format that aligns with LLM caption embeddings has potential computational advantages beyond being information-rich, contextual and embedded in world knowledge. Indeed, such rich representations may act as a suitable candidate for communication between different brain systems: if for example both visual and auditory processing project to a common (LLM-like) space, information from these modalities can easily be combined and used by other brain processes. Given that LLMs have been shown to be good models for predicting brain activities in language areas too[104,105], another benefit of this code would be that it may allow for easy communication with other organisms[106]. Altogether, our results suggest that large language models provide a general representational format to account for complex information extracted by the brain, and offer exciting novel directions for the field of neuro AI.

## Methods

*The Natural Scenes Dataset*

A detailed description of the Natural Scenes Dataset (NSD; http://naturalscenesdataset.org) can be found in [46]. This dataset contains measurements of fMRI responses from 8 participants who each viewed 9,000–10,000 distinct colour natural scenes over the course of 30–40 scan sessions, comprising a total of 73,000 images, with 3 repetitions per image. Scanning was conducted at 7T using whole-brain, gradient-echo EPI at 1.8-mm isotropic resolution and 1.6-s repetition time. Images were taken from the COCO image dataset and were presented for 3 s with 1-s gaps in between images. A special set of 1,000 images were shared across subjects; the remaining images were unique and mutually exclusive across subjects (note that some participants did not complete 3 trials for each image, therefore, only 515 shared images were seen 3 times by all participants). Subjects fixated centrally and performed a long-term continuous recognition task on the images. The data were pre-processed by performing one temporal interpolation (to correct for slice time differences) and one spatial interpolation (to correct for head motion) and then using a general linear model to estimate single-trial beta weights. In this paper, we used the 1.8-mm volume preparation of the NSD data (betas_fithrf_GLMdenoise_RR).

*LLM embeddings for NSD stimuli*

Captions describing the content of each natural scene were obtained from 5 human observers as part of the COCO dataset. For each NSD participant and for each image presented to the participant, we gathered the five captions provided for that image and took the mean of the resulting embeddings (to account for inter-rater differences). In detail, each of the five captions was passed through an LLM, and we take the average embedding across the captions. For MPNet, we used the all-mpnet-base-v2 version (https://www.sbert.net/docs/pretrained_models.html). Note that this version of MPNet was finetuned to have consistent embeddings for different sentences describing the same scene on COCO (on which NSD is based) and other datasets. This ensures that captions written by different people project to a similar point in embedding space, amplifying the ability of the model to extract cross-observer, consistent semantic meaning from captions in the NSD dataset.

In Fig. 2A, we also retrieved the COCO category words for each image (i.e., the words associated with the COCO category labels present in the image), concatenated these category words into a string and fed this string into the LLM (called *LLM* in the figure). In Fig. 2B, we did the same for all nouns and verbs of the captions (using the Natural Language Toolkit (nltk) python library[107] to determine which words were nouns/verbs; respectively called *LLM nouns* and *LLM verbs* in the figure). In Fig. 2C, we also used a single-word-wise LLM embedding (called *LLM* in the figure). Here, we fed each word from each of the 5 COCO captions of each image separately into the LLM, and retrieved the average embedding (similarly to how one would use single word embeddings). Finally, in Supp. Fig. 5, we compared several different LLMs using our standard approach of averaging their embeddings across the 5 COCO captions.

*Category labels for NSD stimuli*

For the multi-hot control in Fig. 2, as well as for training our category-trained ANNs, we used multi-hot binary vectors based on the category labels provided by COCO for each image (i.e., vectors of 0s with 1s for each category present in the image).

*Word embeddings for NSD stimuli*

For word embedding control models (as opposed to the sentence embeddings described above), we used fasttext[58,59] and GloVe[60]. Using the same COCO image captions as above, we constructed several distinct models, each capturing different aspects of the captions. Word embeddings can be combined additively (a standard example is "queen"="king"-"man"+"woman"), and so we average the embeddings across words. In Fig. 2A, for category word embeddings, we averaged the word embeddings for each COCO category label. In Fig. 2C, we combined the embeddings for all words in the scene captions by taking the mean embedding across all words of all five COCO captions. Some words were not recognised by fasttext or GloVe because they were either misspelt or did not exist in the corpus. For these cases, we either corrected the misspelling, found a similar word in the fasttext corpus, or removed them. In rare cases, a stimulus may have no category words. In these cases, we used the word embedding for "something" (this is done because every stimulus needs an embedding for RSA, and "something" is a neutral term).

*Artificial Neural Network activations for NSD stimuli*

For all ANNs, we collect activities for the layer (and timestep in the case of RCNNs) of interest for all NSD images. We pre-process stimuli to match the input range expected by each model.

*Quantifying model-brain representational agreement using Representational Similarity Analysis*

We used representational similarity analysis to quantify the match between various models described above and brain representations on the entire NSD dataset. We apply this analysis both ROI-wise (using the 'streams' ROI definitions of NSD) and in a searchlight fashion[108,109].

Representational dissimilarity matrices (RDMs) were constructed from participants' native space single-trial beta weights. Analyses were restricted to images that had been seen three times by the participant, and beta weights were *z*-scored across single trials within each scanning session for each participant. We then averaged over each image's three repetitions to get an average response estimate for each image. In the searchlight analysis, for each voxel *v*, we extracted activity patterns in a sphere centred at *v* with a radius of 6 voxels (only keeping spheres with more than 50% voxels inside the brain; when a sphere included voxels outside the brain, these voxels were excluded from the analysis). Activity patterns were compared between pairs of stimuli using Pearson correlation distances to create RDMs.

Given the large scale of the NSD dataset, to relate the brain RDMs to the model RDMs, we devised a practical sampling procedure based on independent subsets of images. We first randomly sampled 100 NSD stimuli from the participant's 10,000 images. We indexed the brain activity patterns for these 100 images and constructed the RDM for this subset. We also indexed the model RDMs to retrieve the pairwise distances for the same 100 stimulus images. This led to 100 x 100 symmetric RDMs, with an upper-triangular vector length of 4,950 pairs (one for each model/RCNN layer & timestep, and one for each ROI/searchlight sphere). These upper-triangular RDMs were then compared between brain and model using Pearson correlation in each ROI/searchlight sphere. There was one such correlation per ROI for each participant-model comparison. The randomly sampled 100 images were then removed from the image sampling pool, and we repeated the sampling procedure until we had exhausted all 10,000 images. This resulted in 100 independent correlation volumes, which were averaged. (Note: 4 participants completed the full NSD experiment, while another two had seen all three repetitions of 6,234 images and 2 participants had seen the three repetitions of 5,445 images, leading to 100 splits, 62 splits, or 54 splits depending on the participant).

For the ROI analyses, each subject's result was noise corrected. The subject-wise noise ceiling was approximated as the correlation between this subject's RDM and the mean RDM across all other subjects (these RDMs were computed on the shared 515 NSD images seen by all participants). Intuitively, this can be seen as pitting the model against the average of 7 human subjects: if the model predicts the subject's data as well as the mean of 7 humans, it has reached the noise ceiling. These subject-wise noise-ceiling corrected correlations were then averaged. Significance was tested using 2-tailed *t*-tests across the 8 NSD participants, and corrected for multiple comparisons using the Benjamini & Hochberg [110] procedure for controlling the false discovery rate with $p = 0.05$. For model comparisons, we tested the significance of the difference between model correlations against 0.

For the searchlight analyses, group-level statistics reported in the manuscript are performed using 2-tailed *t*-tests across the 8 NSD participants, and corrected for multiple comparisons using the Benjamini & Hochberg procedure for controlling the false discovery rate with $p = 0.05$. In the case of individual model maps, we tested the model's correlation against 0. In the case of model comparisons, we tested the significance of the difference between model correlations against 0. Average correlation maps participants, thresholded with our group-level statistics are then projected in freesurfer's fsaverage surface space, and visualised on a flattened cortical flatmap.

*Encoding model*

We trained a linear encoding model to predict voxel activities from MPNet embeddings (Fig. 1C). We apply this analysis to the full brain. We used a regularised linear regression framework that was solved for each subject separately. In this framework, the modelled data, **y**, consists of the brain activity measurements (*n* images x *p* voxels) and the predictors, **X**, consist of MPNet embeddings for each image (*n* images x 768 MPNet_dimensions).

We set aside the shared 515 NSD images seen 3 times by all participants as a test set. We used fractional ridge regression[54] to estimate the parameters, $\hat{h}$ (*p* voxels x MPNet_dimensions) for 20 different regularisation fractions (0.05 to 1 in increments of 0.05), using 5-fold cross-validation. The fraction that best predicted each embedding feature after cross-validation was identified, and used as the final model. To evaluate the model, we computed the Pearson correlation for each voxel between the predicted activities and the true activities on the test set. The group-level statistics reported in the manuscript are performed using 2-tailed *t*-tests across the 8 NSD participants, and corrected for multiple comparisons using the Benjamini & Hochberg procedure for controlling the false discovery rate with $p = 0.05$.

*Encoding model based brain activity predictions*

Our encoding model allows us to predict the brain activities from any sentence. That is, we can predict the activities that would be evoked if the participant saw an image captioned by that sentence. To this end, we simply write a sentence, project it in LLM embedding space, and use the resulting embedding as input to our encoding model. To test this approach, we reproduced contrasts from the neuroscientific literature (Fig. 1E). In each contrast, we write 5 sentences for each group, we average the predicted activities, and plot the contrast between these activities on brain maps (unlike all other maps in this paper, there is no correction for false discovery rate). We did not have a precise method for selecting these sentences, and simply attempted to make them representative of the contrasts we aimed to reproduce. The sentences we used for each contrast were:

*People*: 'Man with a beard smiling at the camera.', 'Some children playing.', 'Her face was beautiful.', 'Woman and her daughter playing.', 'Close up of a face of young boy.'

*Places*: 'A view of a beautiful landscape.', 'Houses along a street.', 'City skyline with blue sky.', 'Woodlands in the morning.', 'A park with bushes and trees in the distance.'],

*Food*: 'A plate of food with vegetables.', 'A hamburger with fries.', 'A bowl of fruit.', 'A plate of spaghetti.', 'A bowl of soup.'

*Decoding of LLM embeddings from brain data*

We decoded captions from visually evoked activity by learning a linear mapping from brain activity to the LLM caption embeddings (this mapping can be seen as the inverse mapping to the encoding model described above), followed by a dictionary lookup scheme[55] (Fig. 1F).

We apply this analysis on all voxels inside the 'streams' visual ROIs (provided by NSD). We used a regularised linear regression framework that was solved for each subject separately. In this framework, the modelled data, **y**, consists of the captions embeddings (*n* images x 768 MPNet_dimensions) and the predictors, **X**, consists of brain activity measurements (*n* images x *p* voxels).

We set aside a test set to test the performance of the decoding, by holding out the shared 515 NSD images seen 3 times by all participants. We used fractional ridge regression[54] to estimate the parameters, $\hat{\boldsymbol{h}}$ ($p$ voxels x 768 features), that represent the optimal sets of weights to apply to the predictors ($\boldsymbol{X}_{train}$) to best predict each of the captions embedding features ($\boldsymbol{y}$). Specifically, weights were estimated for 20 different regularisation fractions (0.05 to 1 in increments of 0.05), using 5-fold cross-validation. The fraction that best predicted each embedding feature after cross-validation was identified, and the resulting model was evaluated on the test set by using the corresponding weights to predict the captions embeddings.

To quantify the accuracy of our test predictions, we compared the Pearson correlation between the predicted embedding and the target test embedding, and plotted a subject-wise kernel density estimate of these correlations. As a noise ceiling, we use the internal consistency of the 5 human-generated captions in COCO. To this end, we compute the Pearson correlation between the LLM embeddings of each of the 5 captions and the averaged embedding of the 4 others, and average the resulting 5 correlations.

To obtain a caption reconstruction, we used a simple dictionary lookup scheme. We took the 3.1 million captions from the Google conceptual captions dataset [56] and embedded these captions using MPNet, yielding a lookup dictionary **D** with dimensionality 3.1M captions x 768 features. For each embedding predicted from the brain data, we computed the Pearson correlation with each of the captions in the dictionary. The caption that was closest to the predicted embedding was chosen as the reconstructed caption.

*Recurrent Convolutional Neural Network (RCNNs)*

Our RCNN models are derived from vNet, a 10-layer convolutional deep neural network architecture designed to closely mirror the progressive increase in foveal receptive field sizes found along the human ventral stream, as estimated by population receptive fields[82]. In contrast to previous instances of vNet, our network is recurrent, including both lateral and top-down recurrent connections following a convolutional pattern, as implemented by Kietzmann et al.[111].

We used the COCO dataset for training. Since the NSD dataset is based on a subset of COCO, we removed the 73,000 images of the NSD dataset from the training and validation sets, and used them as our testing set (i.e., the networks did not see any of the NSD images during training, nor in validation). This resulted in 48,236 COCO images for training, 2,051 for validation, and the 73,000 images part of both COCO and NSD for testing. For rectangle images, we took the largest possible square crop, as was done for the NSD experimental stimuli. Images were resized to 128 x 128 pixels.

We trained our recurrent vNet to map from pixels, i.e. COCO images, to LLM embeddings (i.e. MPNet embeddings of COCO captions extracted as described in *LLM embeddings for NSD stimuli*). The readout layer therefore was 768-dimensional, to match MPNet embeddings (we did not apply a traditional non-linearity softmax or sigmoid activation function to the readout, as MPNet embeddings can be both positive and negative). The objective of the network was to minimise the cosine distance between the predicted and the target LLM embedding. To account for possible variation due to the network randomly initialised parameters, we trained 10 instances with different random seeds[68].

As a stringently controlled comparison model, we trained a separate vNet with identical architecture on a category objective (i.e. minimising cosine distance using a multi-hot encoding of the category labels provided in the COCO dataset for each image, this model has a sigmoid activation function, as is usual for multiclass categorization). Again, we trained 10 instances with different random seeds.

To show that the advantage of training on LLM embeddings is not restricted to this current RCNN architecture, we reproduced these results using a ResNet50[70] architecture instead of our RCNNs (one seed each). We used non-pretrained ResNet50, which we trained to predict either LLM embeddings or category labels, as we did for our RCNNs.

All networks were trained using an Adam optimizer with a learning rate of 5e-2 and an epsilon of 1e-1 for 200 epochs, with a warm up phase of 10 epochs where the learning rate was linearly increased, followed by a cosine decay. We used a batch size of 96 for RCNNs and 512 for ResNets.

Code for all trained networks will be made available with the journal version of this paper.

*RCNN finetuning*

To test if category labels (resp. LLM embeddings) can be decoded from LLM-trained (resp. category-trained) RCNN activities, we performed finetuning experiments. We collected activities for the last layer and timestep from each of the 10 instances of each network on the entire NSD dataset (collecting activities in this way is equivalent to freezing the weights of the network, but does not require recomputing the activations at each epoch). We used the first 71000 images of nsd as a training set, and set aside the last 2000 as a test set. We trained linear readouts to decode multi-hot category labels (resp. LLM embeddings) from the activities of LLM-trained (resp. category-trained) networks, by minimising the cosine distance between prediction and target (as described above for training the full RCNN; the readout activation, optimizer, and training hyperparameters were also the same as for training the full RCNN). We then average the test performance across the 10 network instances with different seeds. As a noise floor, we computed the mean LLM embedding (resp. multi-hot vector) across the 48,238 images used to train the RCNNs, and computed the mean cosine distance with the LLM embedding (resp. multi-hot vector) of the 2,051 validation images.

*Other ANNs*

We tested several other ANNs. This includes:

Supervised models (object category)

- We trained an RCNN on object classification on the ecoset dataset[63]. To help the network deal with this larger dataset, we doubled the number of channels. Otherwise this network was identical to the previous RCNNs.
- CORnet-S [112] trained on imagenet[77], taken from thingsvision[113].

- Alexnet[114] trained on imagenet, taken from brainscore[76].
- Alexnet-gn trained on imagenet, taken from[65].
- resnet50 trained on imagenet, taken from brainscore.
- Nf-resnet50 trained on imagenet (best performing CNN on predicting NSD data in [75], taken from timm[115].

Supervised models (scene category)

- We trained a ResnNet50 trained on scene categorization on the places365 dataset[78].
- A ResNet50 trained on scene categorization on the taskonomy dataset[79], taken from https://github.com/StanfordVL/taskonomy.

Semi-Supervised models

- Resnext101_32x8d_wsl[80], trained on 914M public images (best brainscore model available to download), taken from https://pytorch.org/hub/facebookresearch_WSL-Images_resnext/.
- CLIP_RN50_imgs (i.e., the visual stream of CLIP with a ResNet50 backbone)[43], trained on webimagetext[43].
- CLIP_ViT (i.e., the visual stream of CLIP with a vision transformer backbone), trained on webimagetext, taken from https://github.com/openai/CLIP.

Unsupervised models

- Alexnet, trained using instance-prototype contrastive learning (IPCL) on imagenet, taken from [65].
- ResNet50, trained using SimCLR[81] on imagenet, taken from https://github.com/google-research/simclr.

*Predicting brain activity from ANN activations*

To compare the representations in our networks to the brain's representations, we apply a similar RSA approach as described above. First, RDMs for all images in the NSD dataset are computed in the layer (and timestep) of interest in the networks. Second, correlations between RCNNs and brain RDMs are computed, ROI-wise or in a searchight fashion. To quantify how well layer L at timestep T predicts brain activity, we computed the Pearson correlation between the RDM for layer L at timestep T and the brain data RDM at each ROI or searchlight location. In the case of our RCNNs, for which we have 10 instances with different random seeds, we compute individual RDMs for each seed, and then average correlations with brain data across seeds.

**Acknowledgements**

The authors acknowledge support by SNF grant n.203018 (Doerig), the ERC stg grant 101039524 TIME (Doerig & Kietzmann), NIH grant R01EY034118 (Kay), an ERC stg grant 759432 START (Charest), a Courtois Chair in computational neuroscience (Charest), and an NSERC Discovery grant (Charest).

Collection of the NSD dataset was supported by NSF IIS-1822683 (Kay) and NSF IIS-1822929 (Naselaris). We thank G. Tuckute and P. Sulewski for helpful discussions.

**Code and data availability**

The Natural Scenes Dataset is available at http://naturalscenesdataset.org. Code for the analyses reported here will be available upon publication of the manuscript.

**References**

1. Kanwisher, N. Functional specificity in the human brain: a window into the functional architecture of the mind. *Proc. Natl. Acad. Sci. U. S. A.* **107**, 11163–11170 (2010).
2. Konkle, T. & Oliva, A. A Real-World Size Organization of Object Responses in Occipitotemporal Cortex. *Neuron* **74**, 1114–1124 (6/2012).
3. Bao, P., She, L., McGill, M. & Tsao, D. Y. A map of object space in primate inferotemporal cortex. *Nature* **583**, 103–108 (2020).
4. Kriegeskorte, N. *et al.* Matching categorical object representations in inferior temporal cortex of man and monkey. *Neuron* **60**, 1126–1141 (2008).
5. Cichy, R. M., Kriegeskorte, N., Jozwik, K. M., van den Bosch, J. J. F. & Charest, I. The spatiotemporal neural dynamics underlying perceived similarity for real-world objects. *Neuroimage* **194**, 12–24 (2019).
6. Kriegeskorte, N. Deep Neural Networks: A New Framework for Modeling Biological Vision and Brain Information Processing. *Annual Review of Vision Science* **1**, 417–446 (2015).
7. Kriegeskorte, N. & Douglas, P. K. Cognitive computational neuroscience. *Nat. Neurosci.* **21**, 1148–1160 (2018).
8. DiCarlo, J. J., Zoccolan, D. & Rust, N. C. How does the brain solve visual object recognition? *Neuron* **73**, 415–434 (2012).
9. Bracci, S. & Op de Beeck, H. P. Understanding Human Object Vision: A Picture Is Worth a Thousand Representations. *Annu. Rev. Psychol.* **74**, 113–135 (2023).
10. Doerig, A. *et al.* The neuroconnectionist research programme. *Nat. Rev. Neurosci.* **24**, 431–450 (2023).
11. Richards, B. A. *et al.* A deep learning framework for neuroscience. *Nat. Neurosci.* **22**, 1761–1770 (2019).
12. Yamins, D. L. K. *et al.* Performance-optimized hierarchical models predict neural responses in higher visual cortex. *Proc. Natl. Acad. Sci. U. S. A.* **111**, 8619–8624 (2014).
13. Khaligh-Razavi, S.-M. & Kriegeskorte, N. Deep supervised, but not unsupervised, models may explain IT cortical representation. *PLoS Comput. Biol.* **10**, e1003915 (2014).
14. Güçlü, U. & van Gerven, M. A. J. Deep Neural Networks Reveal a Gradient in the Complexity of Neural Representations across the Ventral Stream. *J. Neurosci.* **35**, 10005–10014 (2015).
15. Brandman, T. & Peelen, M. V. Interaction between Scene and Object Processing Revealed by Human fMRI and MEG Decoding. *J. Neurosci.* **37**, 7700–7710 (2017).
16. Sadeghi, Z., McClelland, J. L. & Hoffman, P. You shall know an object by the company it keeps: An investigation of semantic representations derived from object co-occurrence in visual scenes. *Neuropsychologia* **76**, 52–61 (2015).
17. Bonner, M. F. & Epstein, R. A. Object representations in the human brain reflect the co-occurrence statistics of vision and language. *Nat. Commun.* **12**, 4081 (2021).
18. Ackerman, C. M. & Courtney, S. M. Spatial relations and spatial locations are dissociated within prefrontal and parietal cortex. *J. Neurophysiol.* **108**, 2419–2429 (2012).
19. Chafee, M. V., Averbeck, B. B. & Crowe, D. A. Representing spatial relationships in posterior

parietal cortex: single neurons code object-referenced position. *Cereb. Cortex* **17**, 2914–2932 (2007).

20. Graumann, M., Ciuffi, C., Dwivedi, K., Roig, G. & Cichy, R. M. The spatiotemporal neural dynamics of object location representations in the human brain. *Nat Hum Behav* (2022) doi:10.1038/s41562-022-01302-0.
21. Zhang, B. & Naya, Y. Medial Prefrontal Cortex Represents the Object-Based Cognitive Map When Remembering an Egocentric Target Location. *Cereb. Cortex* **30**, 5356–5371 (2020).
22. Bar, M. Visual objects in context. *Nat. Rev. Neurosci.* **5**, 617–629 (2004).
23. Russell, B., Torralba, A., Liu, C., Fergus, R. & Freeman, W. Object recognition by scene alignment. *Adv. Neural Inf. Process. Syst.* **20**, (2007).
24. Võ, M. L.-H., Boettcher, S. E. & Draschkow, D. Reading scenes: how scene grammar guides attention and aids perception in real-world environments. *Curr Opin Psychol* **29**, 205–210 (2019).
25. Kaiser, D., Quek, G. L., Cichy, R. M. & Peelen, M. V. Object Vision in a Structured World. *Trends Cogn. Sci.* **23**, 672–685 (2019).
26. Võ, M. L.-H. The meaning and structure of scenes. *Vision Res.* **181**, 10–20 (2021).
27. Epstein, R. A. & Baker, C. I. Scene Perception in the Human Brain. *Annu Rev Vis Sci* **5**, 373–397 (2019).
28. Bartnik, C. G. & Groen, I. I. A. Visual Perception in the Human Brain: How the Brain Perceives and Understands Real-World Scenes. in *Oxford Research Encyclopedia of Neuroscience* (2023).
29. Epstein, R. A. & Kanwisher, N. A cortical representation of the local visual environment. *Nature* **392**, 598–601 (1998).
30. Epstein, R., Harris, A., Stanley, D. & Kanwisher, N. The parahippocampal place area: recognition, navigation, or encoding? *Neuron* **23**, 115–125 (1999).
31. Epstein, R. A. Parahippocampal and retrosplenial contributions to human spatial navigation. *Trends Cogn. Sci.* **12**, 388–396 (2008).
32. Groen, I. I. A., Ghebreab, S., Prins, H., Lamme, V. A. F. & Scholte, H. S. From image statistics to scene gist: evoked neural activity reveals transition from low-level natural image structure to scene category. *J. Neurosci.* **33**, 18814–18824 (2013).
33. Stansbury, D. E., Naselaris, T. & Gallant, J. L. Natural scene statistics account for the representation of scene categories in human visual cortex. *Neuron* **79**, 1025–1034 (2013).
34. Groen, I. I. *et al.* Distinct contributions of functional and deep neural network features to representational similarity of scenes in human brain and behavior. *Elife* **7**, (2018).
35. Brown, T. B. *et al.* Language Models are Few-Shot Learners. *arXiv [cs.CL]* (2020).
36. Cer, D. *et al.* Universal Sentence Encoder for English. in *Proceedings of the 2018 Conference on Empirical Methods in Natural Language Processing: System Demonstrations* 169–174 (Association for Computational Linguistics, Stroudsburg, PA, USA, 2018).
37. Devlin, J., Chang, M.-W., Lee, K. & Toutanova, K. BERT: Pre-training of Deep Bidirectional Transformers for Language Understanding. *arXiv [cs.CL]* (2018).
38. Arora, S., Liang, Y. & Ma, T. A simple but tough-to-beat baseline for sen-tence embeddings. in *International Conference on Learning Representations* (2017).
39. Song, K., Tan, X., Qin, T., Lu, J. & Liu, T.-Y. MPNet: Masked and permuted pre-training for language understanding. *arXiv [cs.CL]* (2020).
40. Lu, J., Batra, D., Parikh, D. & Lee, S. ViLBERT: Pretraining Task-Agnostic Visiolinguistic Representations for Vision-and-Language Tasks. *arXiv [cs.CV]* (2019).
41. Tan, H. & Bansal, M. LXMERT: Learning Cross-Modality Encoder Representations from Transformers. *arXiv [cs.CL]* (2019).
42. Pramanick, S. *et al.* VoLTA: Vision-Language Transformer with Weakly-Supervised Local-Feature Alignment. *arXiv [cs.CV]* (2022).
43. Radford, A. *et al.* Learning Transferable Visual Models From Natural Language Supervision. *arXiv [cs.CV]* (2021).

44. Du, Y., Liu, Z., Li, J. & Zhao, W. X. A Survey of Vision-Language Pre-Trained Models. *arXiv [cs.CV]* (2022).
45. Chen, F.-L. *et al.* VLP: A Survey on Vision-language Pre-training. *Machine Intelligence Research* **20**, 38–56 (2023).
46. Allen, E. J. *et al.* A massive 7T fMRI dataset to bridge cognitive neuroscience and artificial intelligence. *Nat. Neurosci.* **25**, 116–126 (2022).
47. Lin, T.-Y. *et al.* Microsoft COCO: Common Objects in Context. *arXiv [cs.CV]* (2014).
48. Chen, X. *et al.* Microsoft COCO Captions: Data Collection and Evaluation Server. *arXiv [cs.CV]* (2015).
49. Vaswani, A. *et al.* Attention is All you Need. *Adv. Neural Inf. Process. Syst.* **30**, (2017).
50. Cer, D., Diab, M., Agirre, E., Lopez-Gazpio, I. & Specia, L. SemEval-2017 Task 1: Semantic Textual Similarity Multilingual and Crosslingual Focused Evaluation. in *Proceedings of the 11th International Workshop on Semantic Evaluation (SemEval-2017)* 1–14 (Association for Computational Linguistics, Vancouver, Canada, 2017).
51. Kriegeskorte, N., Mur, M. & Bandettini, P. Representational similarity analysis - connecting the branches of systems neuroscience. *Front. Syst. Neurosci.* **2**, 4 (2008).
52. Kriegeskorte, N. & Kievit, R. A. Representational geometry: integrating cognition, computation, and the brain. *Trends Cogn. Sci.* **17**, 401–412 (8/2013).
53. Nili, H. *et al.* A Toolbox for Representational Similarity Analysis. *PLoS Comput. Biol.* **10**, e1003553 (2014).
54. Rokem, A. & Kay, K. Fractional ridge regression: a fast, interpretable reparameterization of ridge regression. *Gigascience* **9**, (2020).
55. Kay, K. N., Naselaris, T., Prenger, R. J. & Gallant, J. L. Identifying natural images from human brain activity. *Nature* **452**, 352–5 (2008).
56. Sharma, P., Ding, N., Goodman, S. & Soricut, R. Conceptual Captions: A Cleaned, Hypernymed, Image Alt-text Dataset For Automatic Image Captioning. in *Proceedings of the 56th Annual Meeting of the Association for Computational Linguistics (Volume 1: Long Papers)* (eds. Gurevych, I. & Miyao, Y.) 2556–2565 (Association for Computational Linguistics, Melbourne, Australia, 2018).
57. Pennock, I. M. L. *et al.* Color-biased regions in the ventral visual pathway are food selective. *Curr. Biol.* **33**, 134–146.e4 (2023).
58. Bojanowski, P., Grave, E., Joulin, A. & Mikolov, T. Enriching Word Vectors with Subword Information. *arXiv [cs.CL]* (2016).
59. Joulin, A., Grave, E., Bojanowski, P. & Mikolov, T. Bag of Tricks for Efficient Text Classification. *arXiv [cs.CL]* (2016).
60. Pennington, J., Socher, R. & Manning, C. GloVe: Global Vectors for Word Representation. in *Proceedings of the 2014 Conference on Empirical Methods in Natural Language Processing (EMNLP)* (eds. Moschitti, A., Pang, B. & Daelemans, W.) 1532–1543 (Association for Computational Linguistics, Doha, Qatar, 2014).
61. Kaplan, J. *et al.* Scaling Laws for Neural Language Models. *arXiv [cs.LG]* (2020).
62. Hernandez, D., Kaplan, J., Henighan, T. & McCandlish, S. Scaling Laws for Transfer. *arXiv [cs.LG]* (2021).
63. Mehrer, J., Spoerer, C. J., Jones, E. C., Kriegeskorte, N. & Kietzmann, T. C. An ecologically motivated image dataset for deep learning yields better models of human vision. *Proc. Natl. Acad. Sci. U. S. A.* **118**, (2021).
64. Kietzmann, T. C., McClure, P. & Kriegeskorte, N. Deep Neural Networks in Computational Neuroscience. *bioRxiv* 133504 (2018) doi:10.1101/133504.
65. Konkle, T. & Alvarez, G. A. A self-supervised domain-general learning framework for human ventral stream representation. *Nat. Commun.* **13**, 491 (2022).
66. Zhuang, C. *et al.* Unsupervised neural network models of the ventral visual stream. *Proceedings of the National Academy of Sciences* **118**, e2014196118 (2021).

67. Spoerer, C. J., Kietzmann, T. C., Mehrer, J., Charest, I. & Kriegeskorte, N. Recurrent neural networks can explain flexible trading of speed and accuracy in biological vision. *PLoS Comput. Biol.* **16**, e1008215 (2020).
68. Mehrer, J., Spoerer, C. J., Kriegeskorte, N. & Kietzmann, T. C. Individual differences among deep neural network models. *Nat. Commun.* **11**, 5725 (2020).
69. Hong, H., Yamins, D. L. K., Majaj, N. J. & DiCarlo, J. J. Explicit information for category-orthogonal object properties increases along the ventral stream. *Nat. Neurosci.* **19**, 613–622 (2016).
70. He, K., Zhang, X., Ren, S. & Sun, J. Deep Residual Learning for Image Recognition. *arXiv [cs.CV]* (2015).
71. Ungerleider, LG., Mishkin, L. Two cortical visual systems. in *Analysis of visual behavior* (ed. Goodale, M., Ingle, D. J., Mansfield, R. J. W.) (MIT Press, 1982).
72. Goodale, M. A. & Milner, A. D. Separate visual pathways for perception and action. *Trends Neurosci.* **15**, 20–25 (1992).
73. Tanaka, K. Inferotemporal cortex and object vision. *Annu. Rev. Neurosci.* **19**, 109–139 (1996).
74. Ishai, A., Ungerleider, L. G., Martin, A., Schouten, J. L. & Haxby, J. V. Distributed representation of objects in the human ventral visual pathway. *Proc. Natl. Acad. Sci. U. S. A.* **96**, 9379–9384 (1999).
75. Conwell, C., Prince, J. S., Kay, K. N., Alvarez, G. A. & Konkle, T. What can 1.8 billion regressions tell us about the pressures shaping high-level visual representation in brains and machines? *bioRxiv* 2022.03.28.485868 (2023) doi:10.1101/2022.03.28.485868.
76. Schrimpf, M. *et al.* Brain-Score: Which Artificial Neural Network for Object Recognition is most Brain-Like? *bioRxiv* 407007 (2018) doi:10.1101/407007.
77. Russakovsky, O. *et al.* ImageNet Large Scale Visual Recognition Challenge. *arXiv [cs.CV]* (2014).
78. Zhou, B., Lapedriza, A., Khosla, A., Oliva, A. & Torralba, A. Places: A 10 Million Image Database for Scene Recognition. *IEEE Trans. Pattern Anal. Mach. Intell.* **40**, 1452–1464 (2018).
79. Zamir, A. *et al.* Taskonomy: Disentangling Task Transfer Learning. *arXiv [cs.CV]* (2018).
80. Mahajan, D. *et al.* Exploring the limits of weakly supervised pretraining. in *Proceedings of the European conference on computer vision (ECCV)* 181–196 (2018).
81. Chen, T., Kornblith, S., Norouzi, M. & Hinton, G. A Simple Framework for Contrastive Learning of Visual Representations. *arXiv [cs.LG]* (2020).
82. Mehrer, J., Spoerer, C. J., Jones, E. C., Kriegeskorte, N. & Kietzmann, T. C. An ecologically motivated image dataset for deep learning yields better models of human vision. *Proc. Natl. Acad. Sci. U. S. A.* **118**, (2021).
83. Güçlü, U. & van Gerven, M. A. J. Semantic vector space models predict neural responses to complex visual stimuli. *arXiv [q-bio.NC]* (2015).
84. Frisby, S. L., Halai, A. D., Cox, C. R., Lambon Ralph, M. A. & Rogers, T. T. Decoding semantic representations in mind and brain. *Trends Cogn. Sci.* **27**, 258–281 (2023).
85. Greene, M. R., Baldassano, C., Esteva, A., Beck, D. M. & Fei-Fei, L. Visual scenes are categorized by function. *J. Exp. Psychol. Gen.* **145**, 82–94 (2016).
86. Greene, M. R. Statistics of high-level scene context. *Front. Psychol.* **4**, 777 (2013).
87. Bar, M. Visual objects in context. *Nat. Rev. Neurosci.* **5**, 617–629 (2004).
88. Henderson, J. M. & Ferreira, F. Scene Perception for Psycholinguists. in *The interface of language, vision, and action: Eye movements and the visual world , (pp* (ed. Henderson, J. M.) vol. 399 1–58 (Psychology Press, xiv, New York, NY, US, 2004).
89. Greene, M. R. & Oliva, A. The briefest of glances: the time course of natural scene understanding. *Psychol. Sci.* **20**, 464–472 (2009).
90. Malcolm, G. L. & Shomstein, S. Object-based attention in real-world scenes. *J. Exp. Psychol. Gen.* **144**, 257–263 (2015).
91. Biederman, I. Perceiving real-world scenes. *Science* **177**, 77–80 (1972).
92. Greene, M. R. Scene Perception and Understanding. in *Oxford Research Encyclopedia of Psychology* (2023).
93. Potter, M. C. Meaning in visual search. *Science* **187**, 965–966 (1975).

94. Carlson, T. A., Simmons, R. A., Kriegeskorte, N. & Slevc, L. R. The emergence of semantic meaning in the ventral temporal pathway. *J. Cogn. Neurosci.* **26**, 120–131 (2014).
95. Contier, O., Baker, C. I. & Hebart, M. N. Distributed representations of behaviorally-relevant object dimensions in the human visual system. *bioRxiv* (2024) doi:10.1101/2023.08.23.553812.
96. Marblestone, A. H., Wayne, G. & Kording, K. P. Toward an Integration of Deep Learning and Neuroscience. *Front. Comput. Neurosci.* **10**, 94 (2016).
97. Golan, T. *et al.* Deep neural networks are not a single hypothesis but a language for expressing computational hypotheses. *Behav. Brain Sci.* **46**, e392 (2023).
98. Geirhos, R. *et al.* Partial success in closing the gap between human and machine vision. *arXiv [cs.CV]* (2021).
99. Popham, S. F. *et al.* Visual and linguistic semantic representations are aligned at the border of human visual cortex. *Nat. Neurosci.* **24**, 1628–1636 (2021).
100. Wang, A. Y., Kay, K., Naselaris, T., Tarr, M. J. & Wehbe, L. Better models of human high-level visual cortex emerge from natural language supervision with a large and diverse dataset. *Nature Machine Intelligence* **5**, 1415–1426 (2023).
101. Tang, J., Du, M., Vo, V. A., Lal, V. & Huth, A. G. Brain encoding models based on multimodal transformers can transfer across language and vision. *arXiv [cs.CL]* (2023).
102. Kay, K., Bonnen, K., Denison, R. N., Arcaro, M. J. & Barack, D. L. Tasks and their role in visual neuroscience. *Neuron* **111**, 1697–1713 (2023).
103. Çukur, T., Nishimoto, S., Huth, A. G. & Gallant, J. L. Attention during natural vision warps semantic representation across the human brain. *Nat. Neurosci.* **16**, 763–770 (2013).
104. Goldstein, A. *et al.* Alignment of brain embeddings and artificial contextual embeddings in natural language points to common geometric patterns. *Nat. Commun.* **15**, 2768 (2024).
105. Schrimpf, M. *et al.* The neural architecture of language: Integrative modeling converges on predictive processing. *Proc. Natl. Acad. Sci. U. S. A.* **118**, (2021).
106. Zada, Z. *et al.* A shared linguistic space for transmitting our thoughts from brain to brain in natural conversations. *bioRxiv* (2023) doi:10.1101/2023.06.27.546708.
107. Bird, S., Klein, E., & Loper, E. *Natural Language Processing with Python: Analyzing Text with the Natural Language Toolkit*. (Reilly Media, 2009).
108. Kriegeskorte, N., Goebel, R. & Bandettini, P. A. Information-based functional brain mapping. *Proc. Natl. Acad. Sci. U. S. A.* **103**, 3863–3868 (2006).
109. Haynes, J. D. & Rees, G. Predicting the stream of consciousness from activity in human visual cortex. *Curr. Biol.* **15**, 1301–7 (2005).
110. Benjamini, Y. & Hochberg, Y. Controlling the False Discovery Rate: A Practical and Powerful Approach to Multiple Testing. *J. R. Stat. Soc. Series B Stat. Methodol.* **57**, 289–300 (1995).
111. Kietzmann, T. C. *et al.* Recurrence is required to capture the representational dynamics of the human visual system. *Proc. Natl. Acad. Sci. U. S. A.* **116**, 21854–21863 (2019).
112. Kubilius, J. *et al.* CORnet: Modeling the Neural Mechanisms of Core Object Recognition. *bioRxiv* 408385 (2018) doi:10.1101/408385.
113. Muttenthaler, L. & Hebart, M. N. THINGSvision: A Python Toolbox for Streamlining the Extraction of Activations From Deep Neural Networks. *Front. Neuroinform.* **15**, 679838 (2021).
114. Krizhevsky, A., Sutskever, I. & Hinton, G. E. ImageNet Classification with Deep Convolutional Neural Networks. in *Advances in Neural Information Processing Systems* (eds. Pereira, F., Burges, C. J., Bottou, L. & Weinberger, K. Q.) vol. 25 (Curran Associates, Inc., 2012).
115. *Timmdocs: Documentation for Ross Wightman's Timm Image Model Library*. (Github).

## Supplementary figures

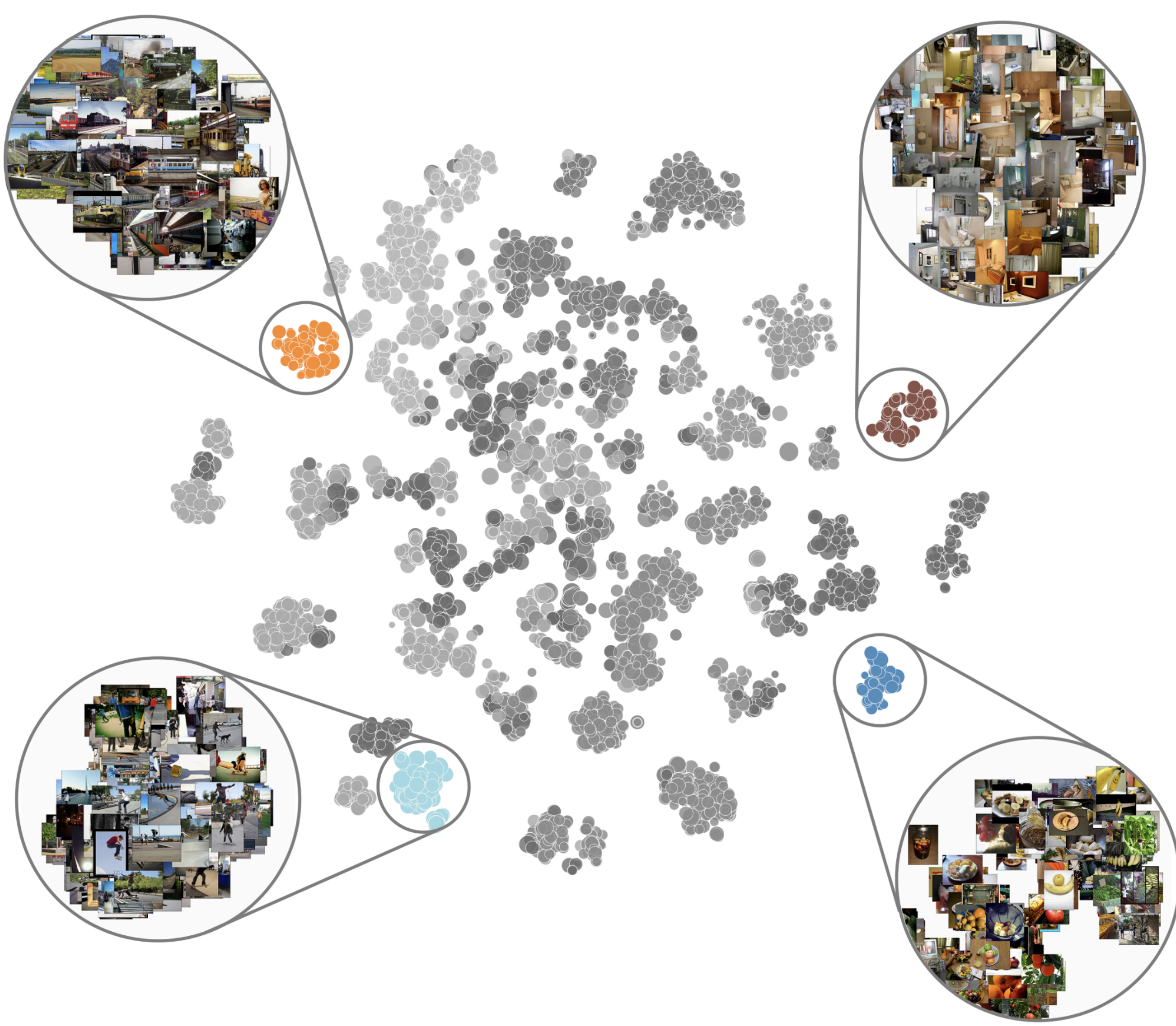

**Supp. Figure 1: Visualising LLM embeddings space for NSD.** 2D T-SNE projection of the MPNet embeddings for the NSD dataset images seen by participant 1. Note that although each element in the dataset is represented here by the associated image for interpretability, T-SNE is computed on MPNet embeddings of the image captions. These embeddings capture relevant semantic information, as shown by the semantically related clusters.

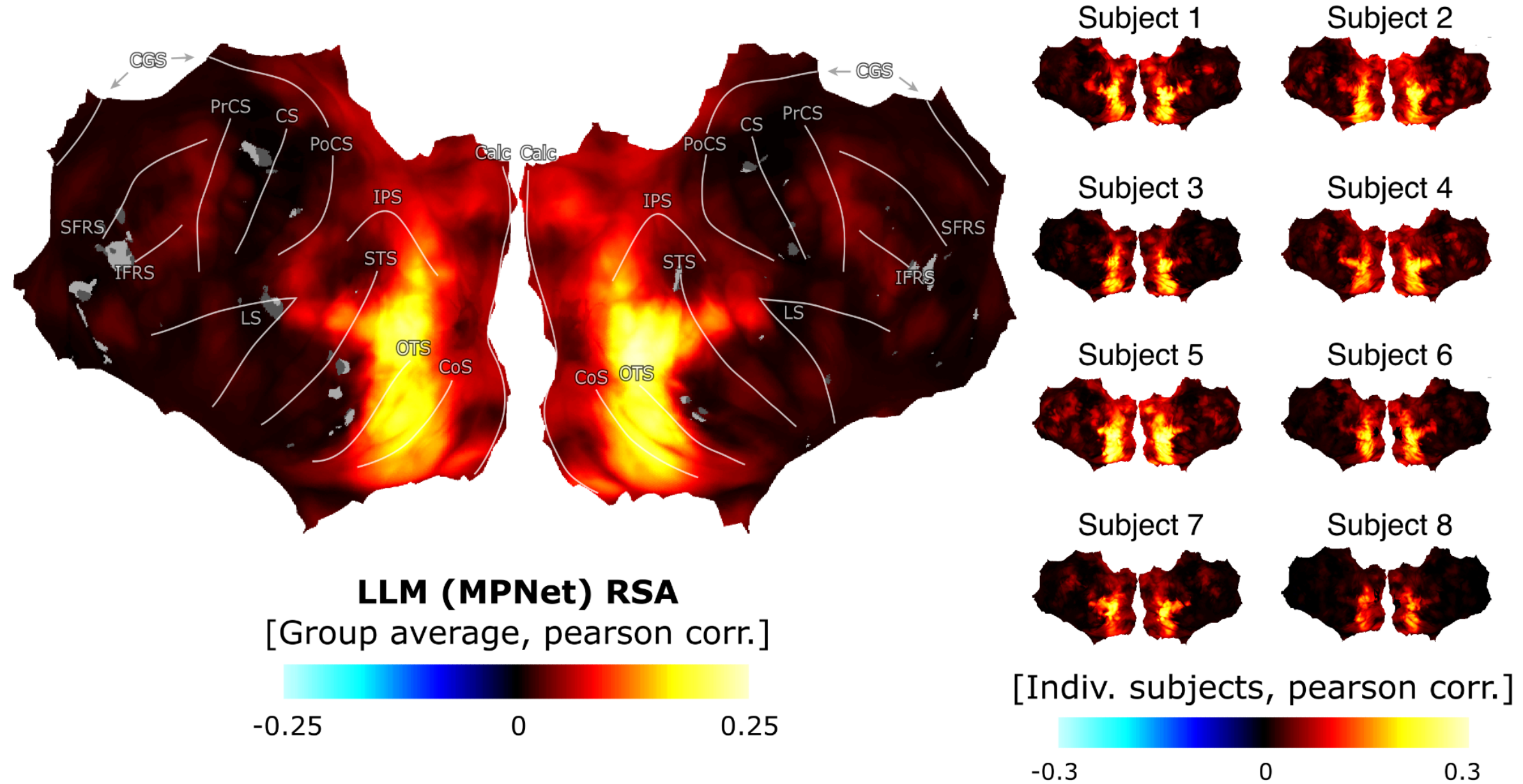

**Supp. Figure 2: MPNet-brain RSA correlation at the group- and subject-level.** This is an extension of Figure 1 B. *Left*: Searchlight map for the correlation between LLM embedding space (as given by MPNet embeddings) and brain representational space. Left: Group average (significance threshold set by a 2-tailed *t*-test across subjects with Benjamini & Hochberg false discovery rate correction; $p$ = 0.05). *Right*: individual subjects, not thresholded for significance.

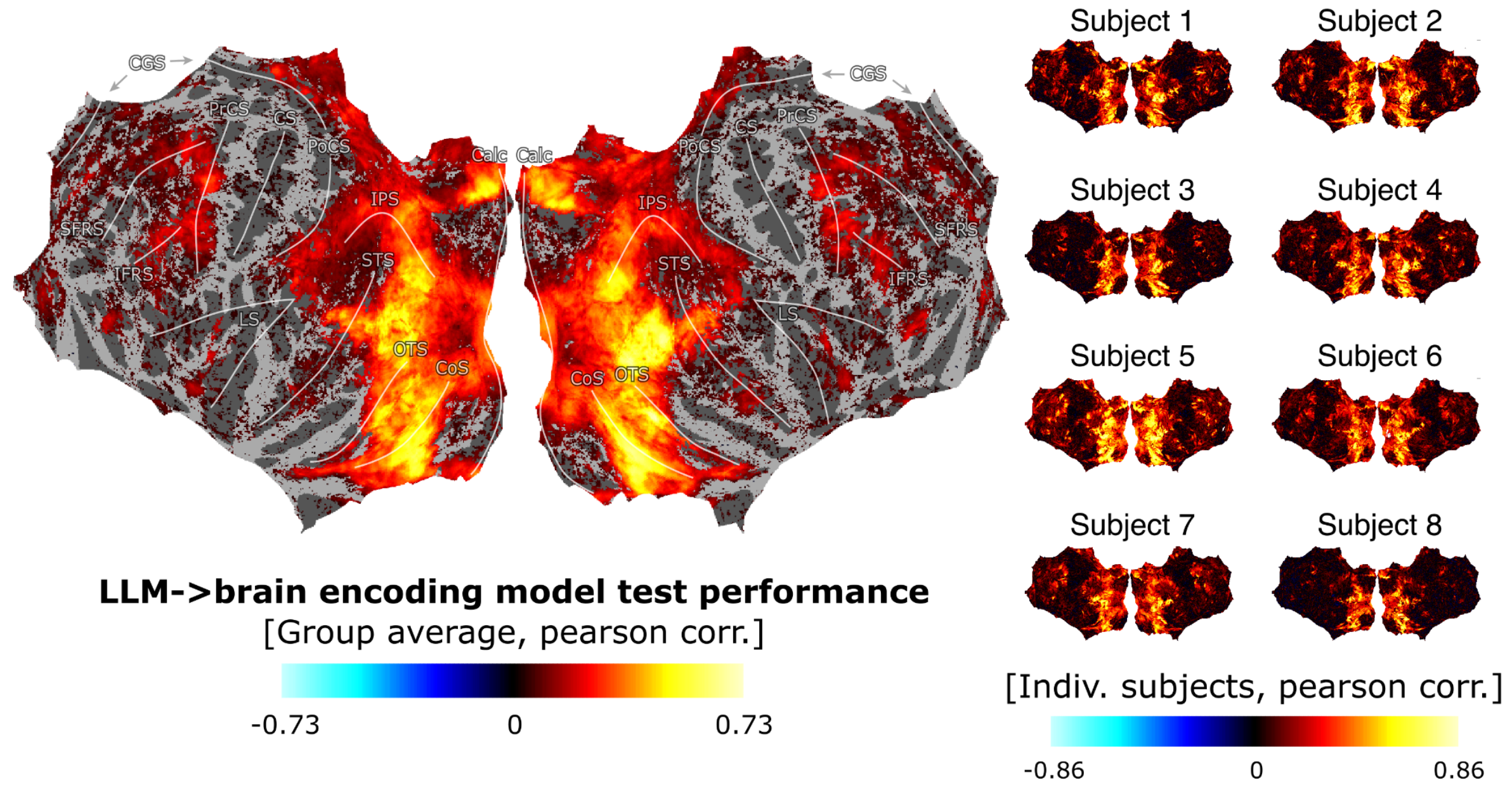

**Supp. Figure 3: MPNet encoding model performance at the group- and subject-level.** This is an extension of Figure 1 C. *Left*: Pearson correlation map between the predicted beta responses on a held out test set, and the actual observed beta responses on this test set, averaged across subjects (significance threshold set by a 2-tailed t-test across subjects with Benjamini & Hochberg false discovery rate correction; p=0.05). *Right*: individual subjects, not thresholded for significance.

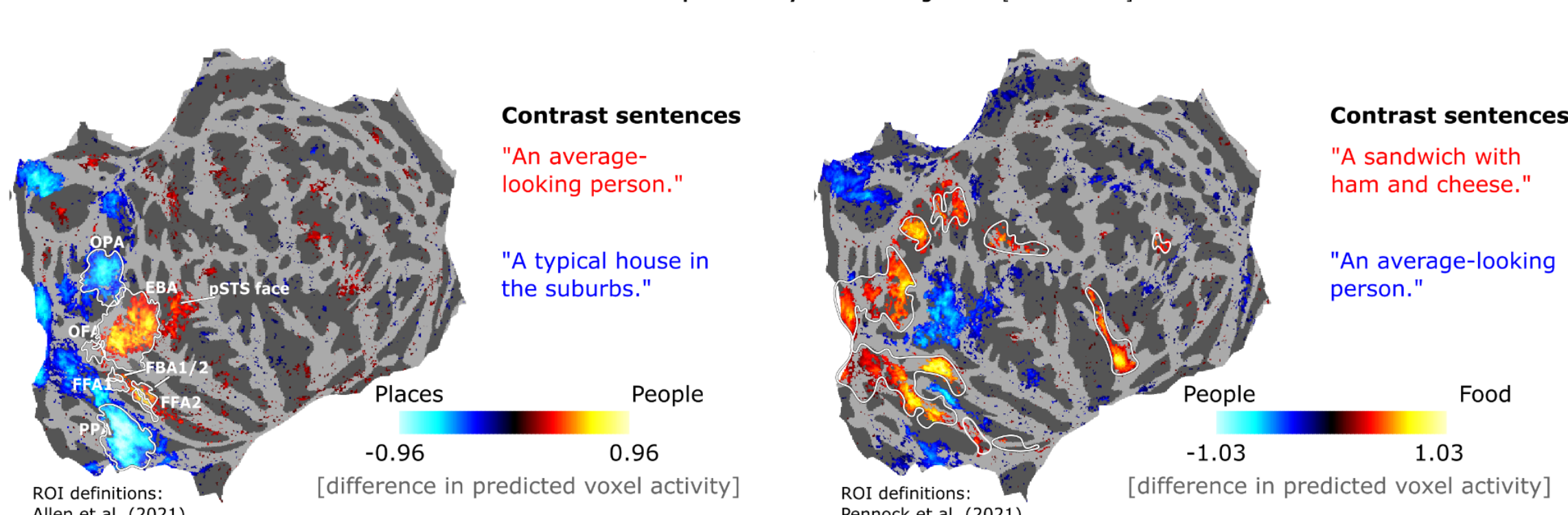

**Supp. Figure 4: Predicting neuroscientific contrasts from single sentences.** We reproduced our results from Fig. 1E using contrasts of brain activities predicted from individual sentences. We used different sentences than the ones from Figure 1E. For each contrast, we write two sentences, obtain the predicted activities, and plot the contrast between these predicted activities on brain maps. The sentences we used for each contrast were 'An average-looking person.' vs. 'A typical house in the suburbs.' for people vs. places (left), and 'A cheese and ham sandwich.' vs. 'An average-looking person.' for food vs. people (right; significance threshold set by a 2-tailed t-test across subjects with $p$=0.05; no correction for false discovery rate was performed). The predicted contrasts reproduce previously described contrasts in the neuroscientific literature (previously described people, place, and food areas are shown as overlays, as in Fig. 1E).

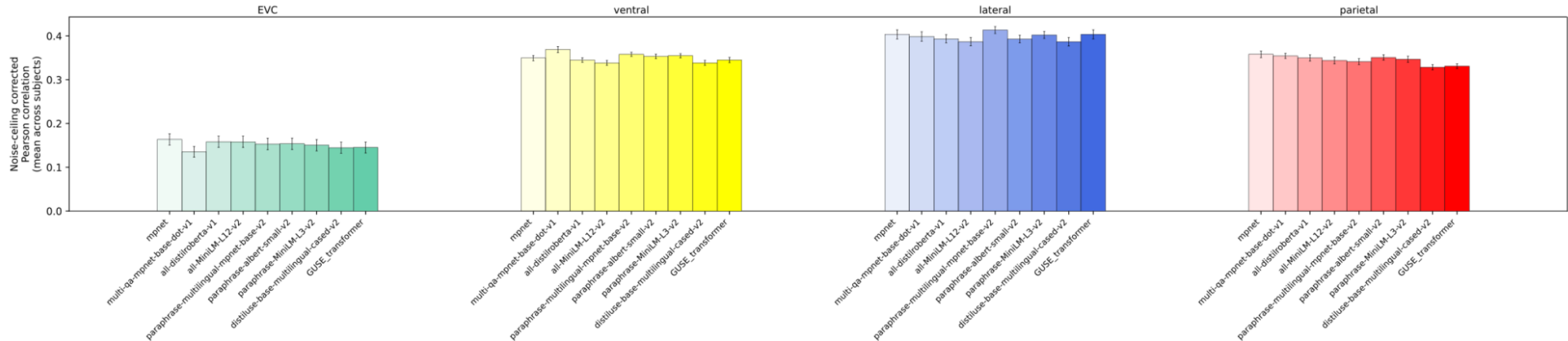

**Supp. Figure 5: Different LLMs all match visually-evoked brain activities similarly well.** To ensure that our results are not reliant on the specific LLM used for embedding the captions, we compared 9 LLMs from the Sentence-Transformers leaderboard (https://www.sbert.net/index.html). We applied our RSA approach in the 'streams' ROI definitions of the NSD dataset, shown in the insert of Fig. 2. The colours of the bars in the plot are colour-coded to match the insert. The x-axis labels denote model names. The match between each model and brain activities is quantified as the noise ceiling corrected correlations between RDMs for each model and a given ROI (error bars reflect standard error across subjects). None of the statistical comparisons among LLM models were found to be significant (2-tailed *t*-test across participants, $p$>0.05, Benjamin & Hochstein false discovery rate corrected). This finding speaks for the generality of our findings, and aligns with previous work indicating that scale can matter more than architectural differences in LLMs[61,62].

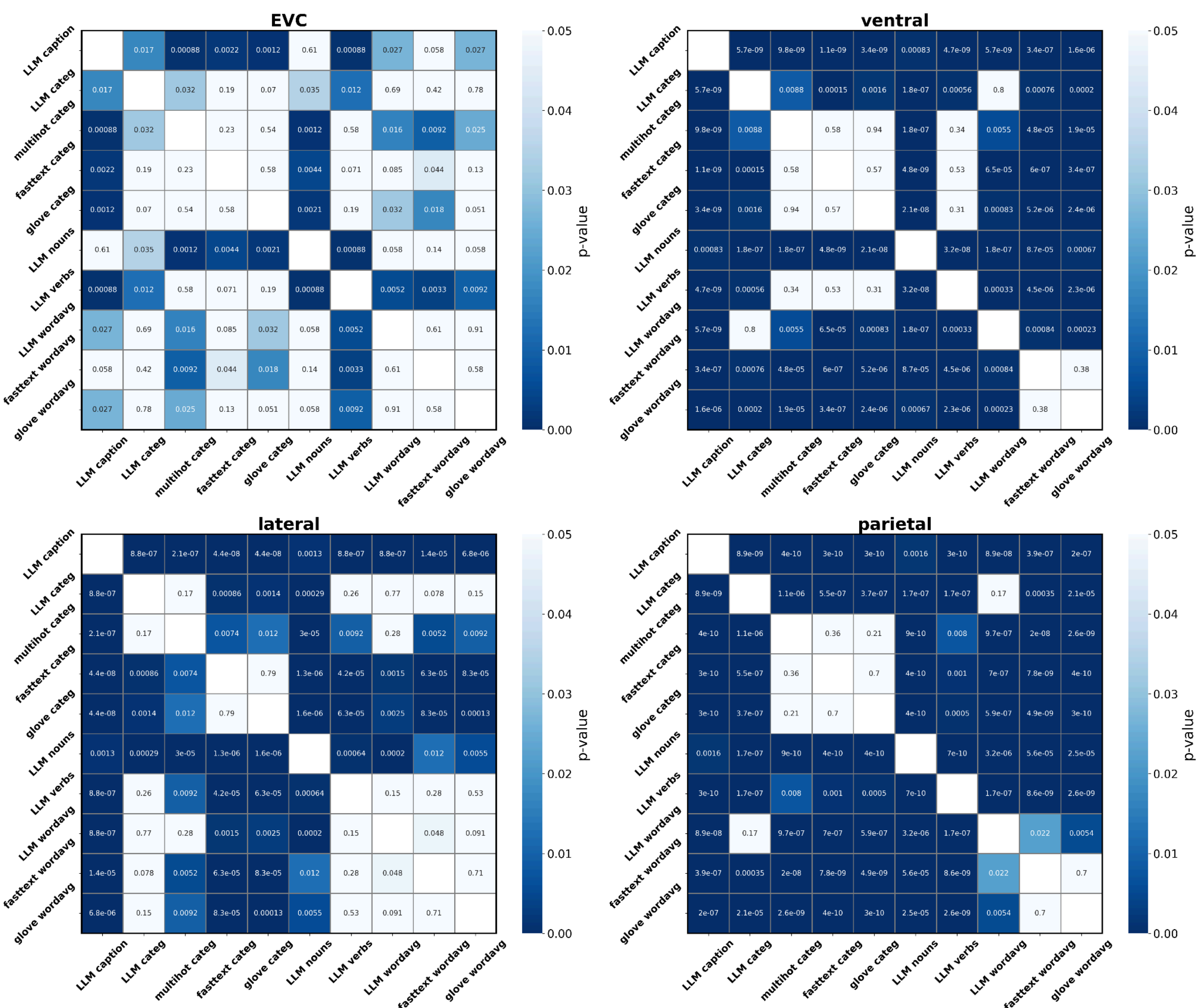

**Supp. Figure 6: All Benjamin & Hochstein false discovery rate corrected p-values for Figure 2.** 2-tailed *t*-test across the 8 NSD participants. Corrections are conducted across all models, independently for each ROI. Model names are listed in the same order as in Fig. 2.

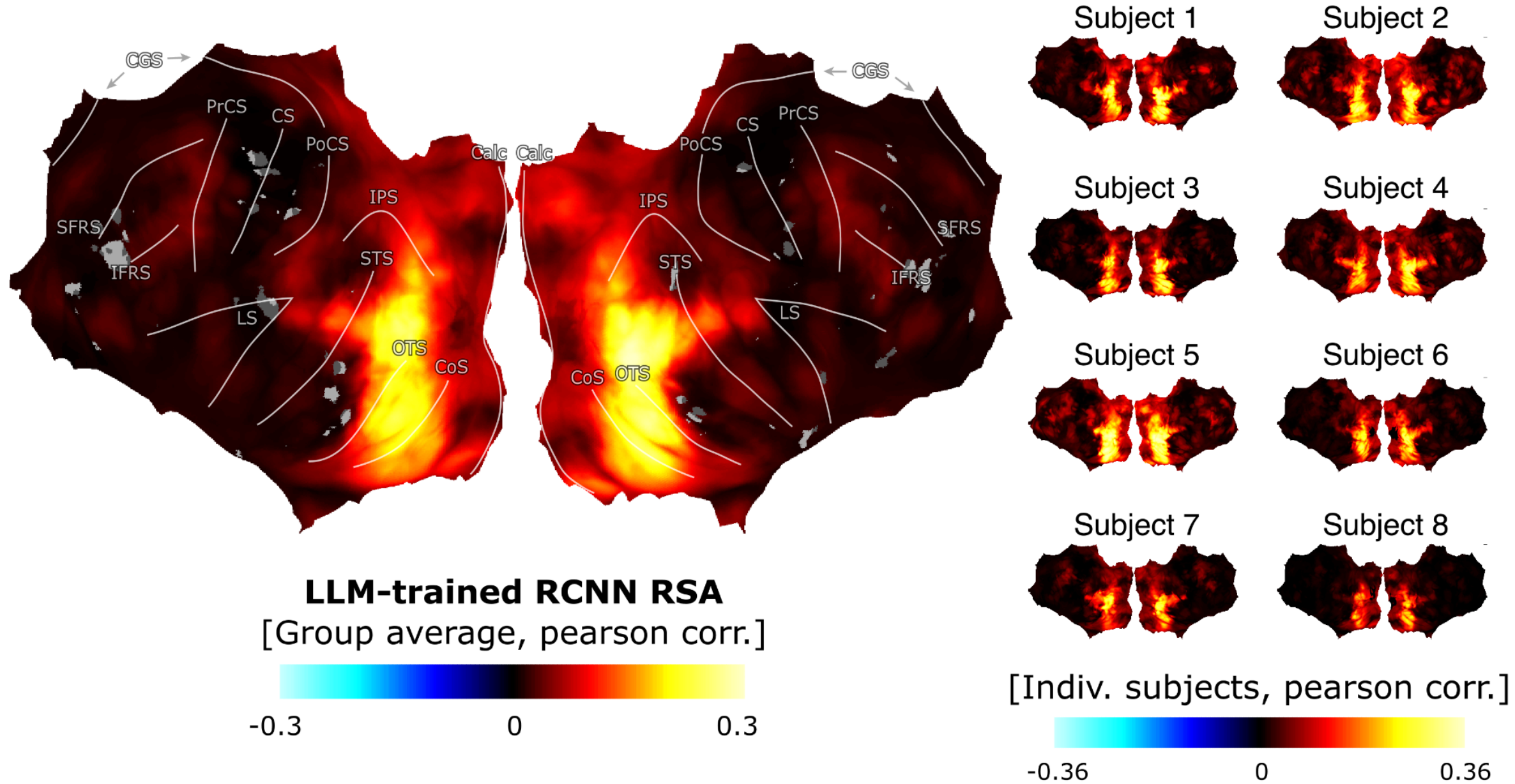

**Supp. Figure 7: *LLM-trained RCNN searchlight* at the group- and subject-level.** *Left*: Searchlight map for the correlation between LLM-trained RCNN representational space (last layer and timestep) and brain representational space. RCNN RDMs are averaged across 10 network instances; correlations are averaged across 8 participants for the group average; significance threshold set by a 2-tailed *t*-test across subjects with Benjamini & Hochberg false discovery rate correction; $p = 0.05$). *Right*: individual subjects, not thresholded for significance

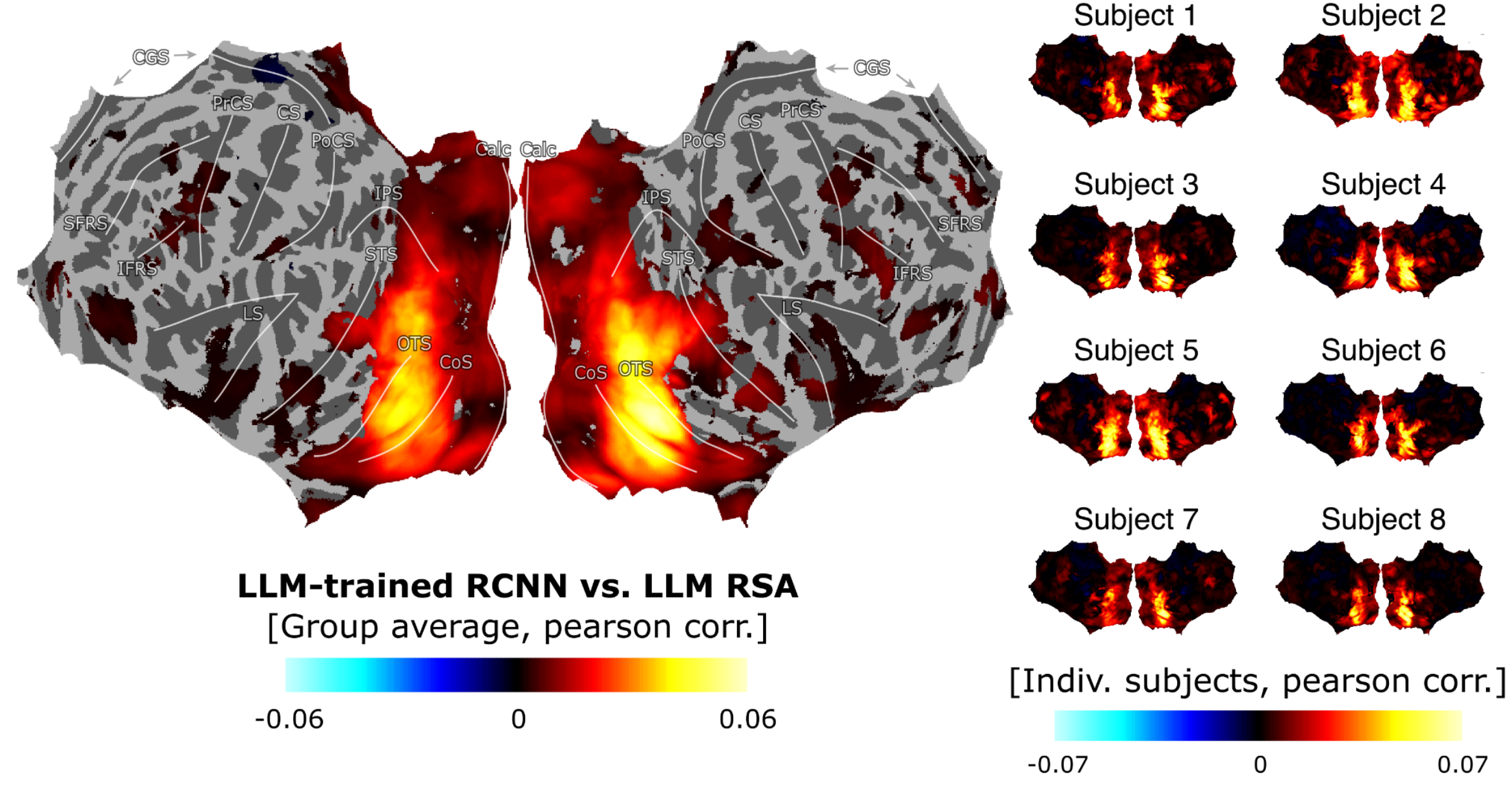

**Supp. Figure 8:** ***LLM-trained RCNN vs. LLM searchlight at the group- and subject-level.*** This is an extension of Figure 3C. *Left*: Searchlight contrast for LLM-trained RCNN (last layer and timestep) vs. LLM embeddings. RCNN RDMs are averaged across 10 network instances; correlations are averaged across 8 participants for the group average; significance threshold set by a 2-tailed *t*-test across subjects with Benjamini & Hochberg false discovery rate correction; $p = 0.05$. *Right*: individual subjects, not thresholded for significance.

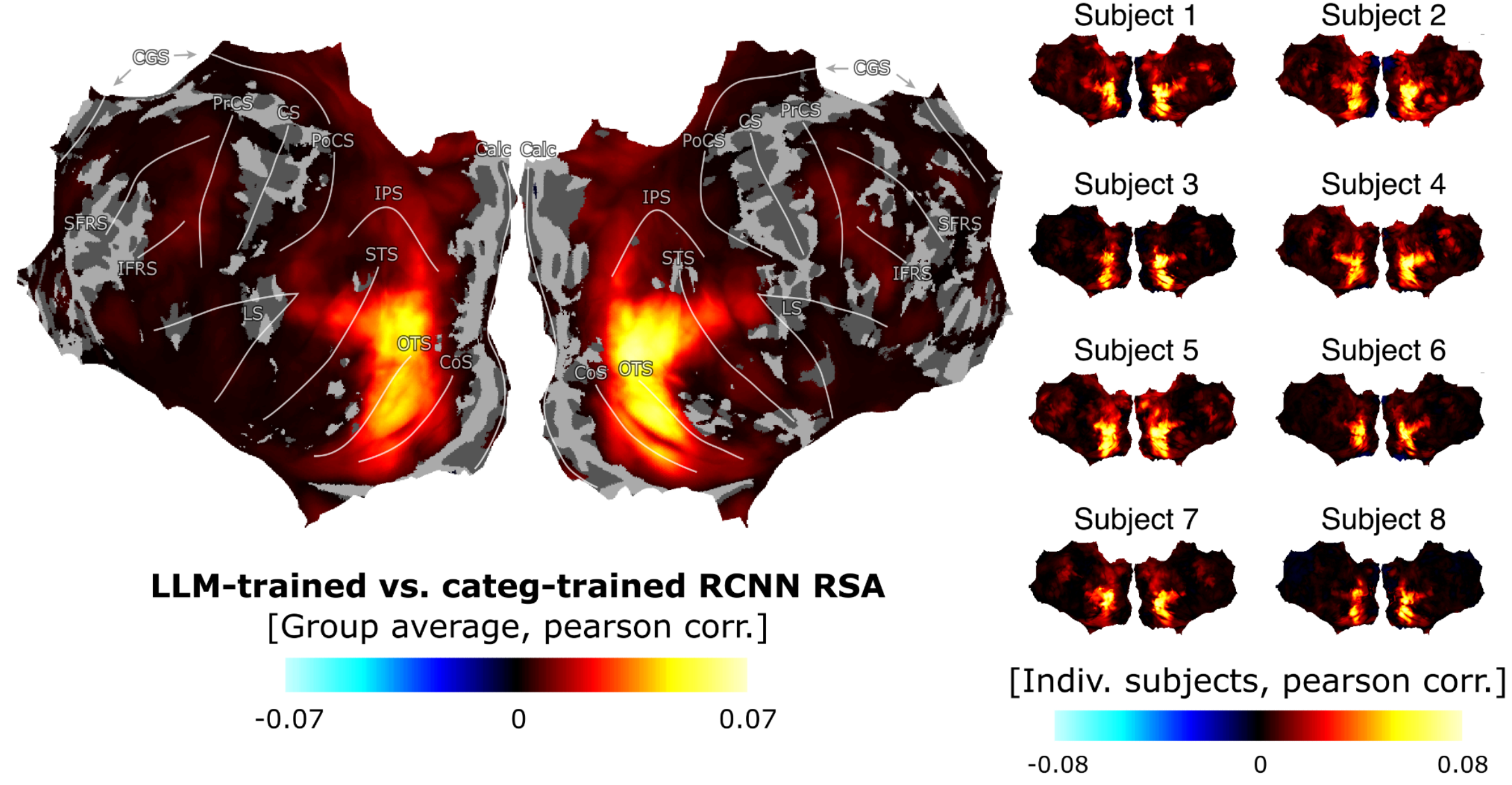

**Supp. Figure 9:** ***LLM-trained RCNN vs. category-trained RCNN searchlight at the group- and subject-level.*** This is an extension of Figure 3D. *Left:* The flatmap shows the group level searchlight contrast (RCNN RDMs are averaged across 10 network instances; correlations are averaged across 8 participants; significance threshold set by a 2-tailed t-test across subjects with Benjamini & Hochberg false discovery rate correction; p = 0.05). *Right*: individual subjects, not thresholded for significance.

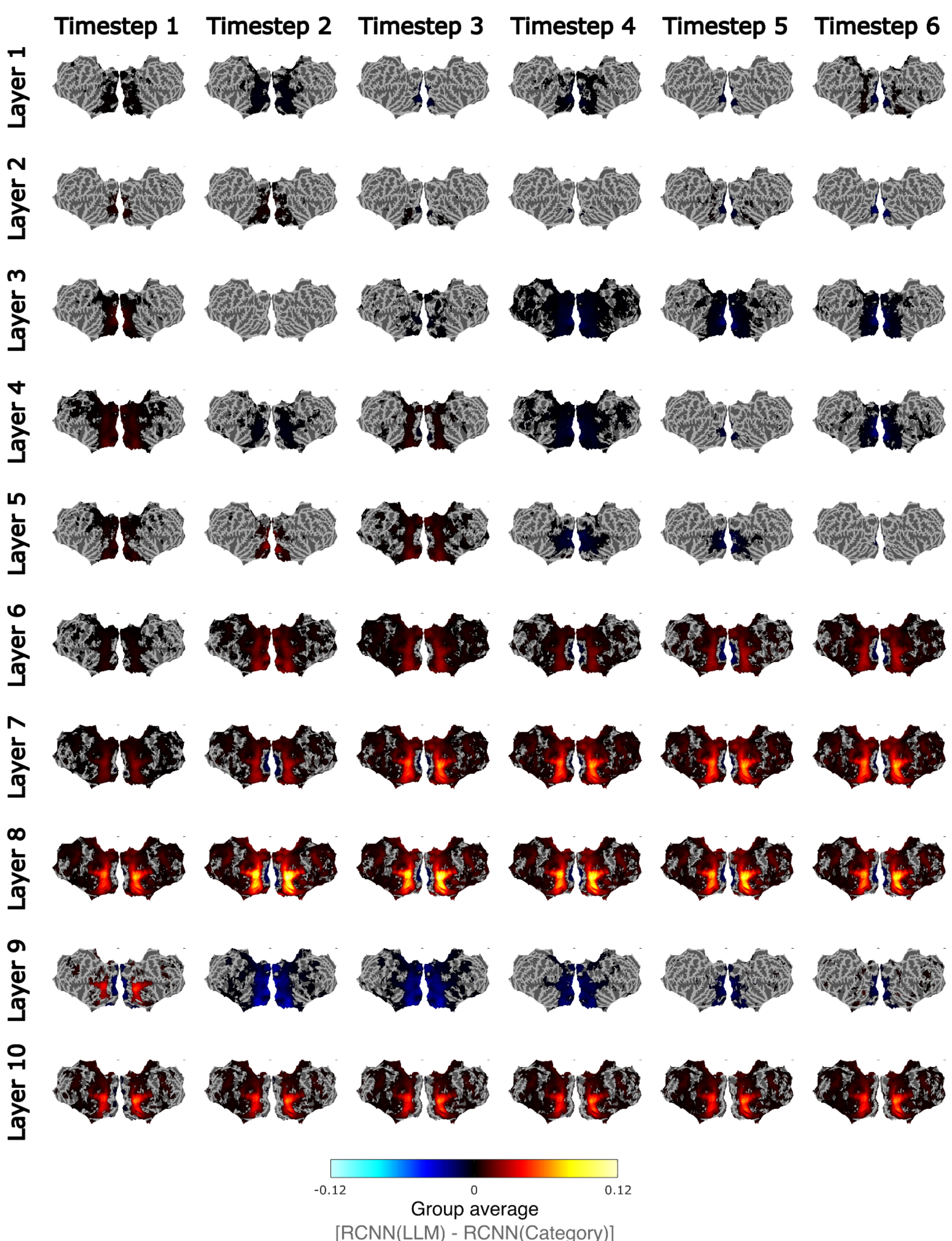

**Supp. Figure 10: LLM-trained vs. category-trained RCNN searchlight contrasts for all layers and timesteps.** Plots show the difference in Pearson correlation with brain data between the LLM-trained and category-trained RCNNs (red/yellow indicate an advantage for the LLM-trained RCNN, and blue indicates an advantage for the category-trained RCNN). Statistics are computed as 2-tailed *t*-tests across subjects, separately for each layer/timestep, and all group level maps are thresholded at $p < 0.05$ with Benjamini & Hochberg false detection rate correction. All flatmaps share the same colormap, shown at the bottom. In later layers (especially 6, 7, 8 & 10), the LLM-trained network has higher representational agreement with higher visual brain areas than the category-trained control. This effect is stronger after recurrence, especially in layer 7.

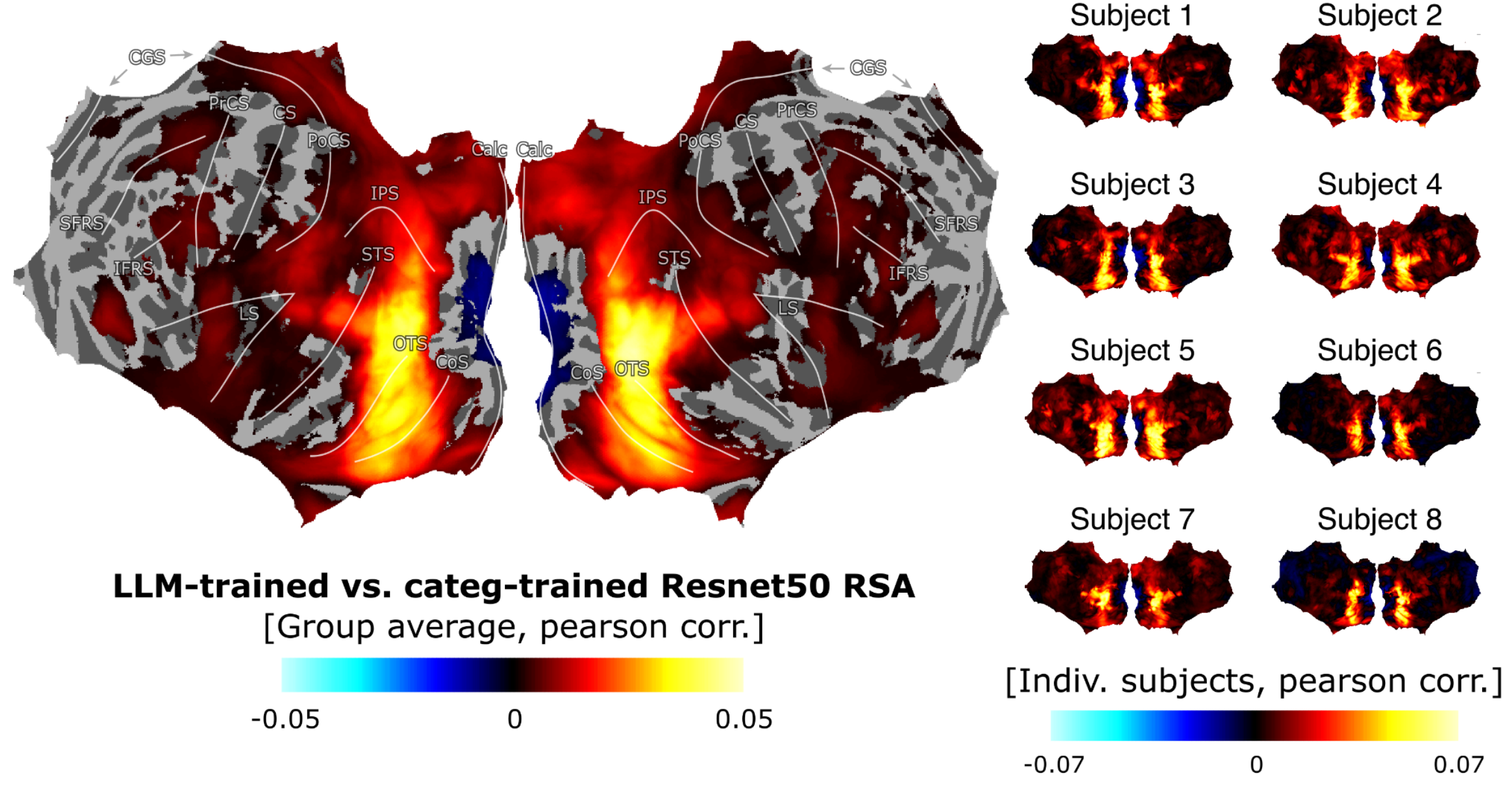

**Supp. Figure 11: Reproducing the benefit in brain alignment of LLM training using ResNet50.** We trained two identical ResNet50s from scratch in the same way as our RCNNs. One predicted LLM embeddings, and the other predicted multi-hot category vectors. *Left*: The flatmap shows the difference in Pearson correlation between the fMRI RDM and the pre-readout layer of these LLM- vs. category-trained ResNet50s at each searchlight location (red/yellow indicate an advantage for the LLM-trained ResNet50, and blue indicates an advantage for the category-trained ResNet50; correlations are averaged across 8 participants; significance threshold set by a 2-tailed t-test across subjects with Benjamini & Hochberg false discovery rate correction; $p$=0.05). The LLM-trained ResNet50 outperforms the category-trained control in a widespread network of higher-level visual areas, reproducing our results obtained using RCNNs. *Right*: individual subjects, not thresholded for significance.

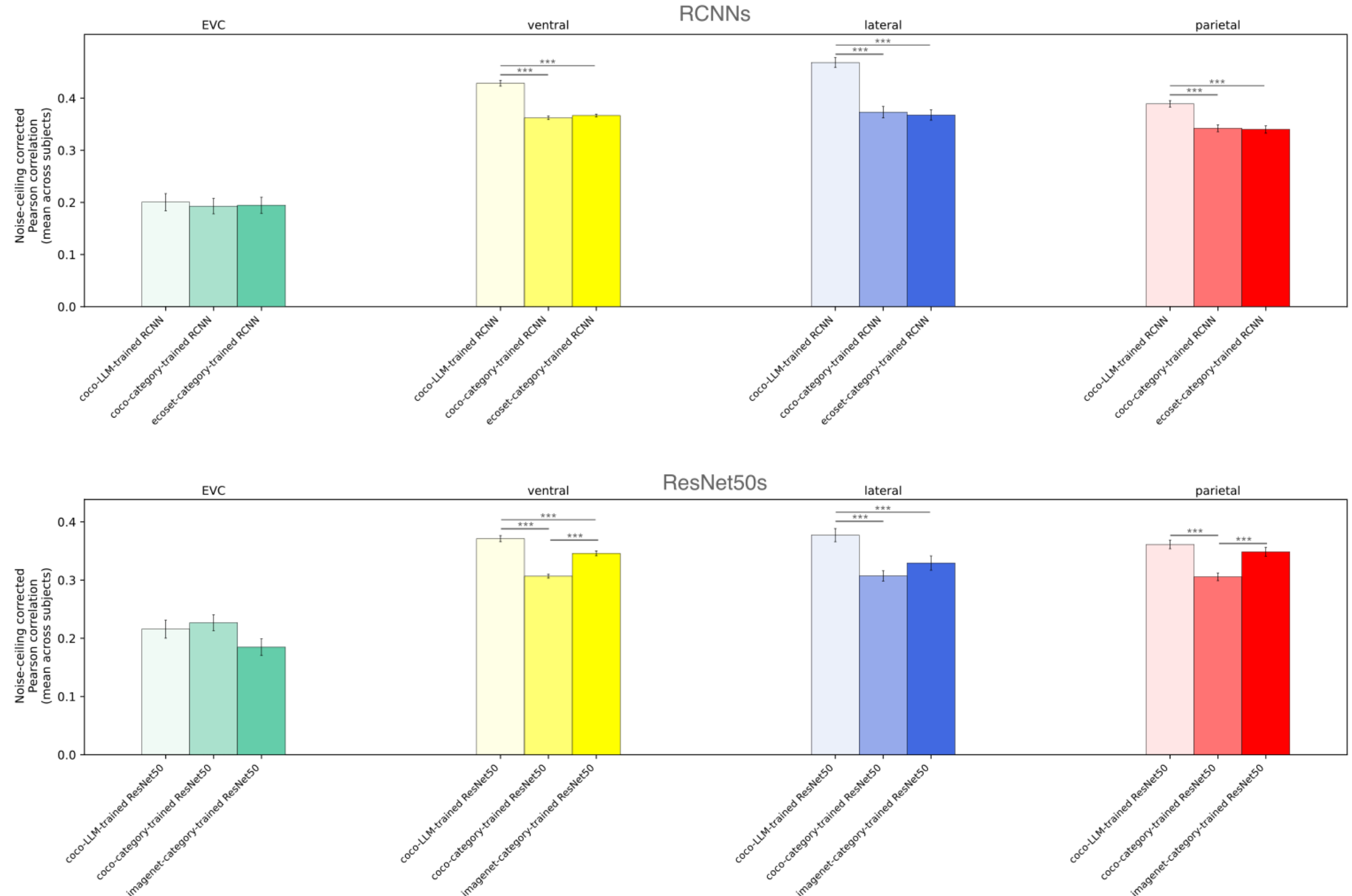

**Supp. Figure 12: The advantage in brain alignment of our LLM-trained models is due to their training objective and not to the dataset.** Our LLM-trained models are trained on the subset of COCO left after removing NSD images. To show that the good brain alignment of our models representations is due to LLM-training and not merely to the training on COCO, we applied the same ROI-wise RSA analysis as in Figs. 2 and 3E, and compared three models (top row): our main RCNN trained on COCO to predict LLM embeddings (coco-LLM-trained RCNN), the control RCNN trained on COCO to predict multi-hot category vectors (coco-category-trained RCNN), and an RCNN trained on ecoset to predict 1-hot category vectors (ecoset-category-trained RCNN; identical to "rcnn_ecoset" model of Fig. 3E). We also conducted the same comparison with ResNet50s instead of RCNNs (swapping ecoset for imagenet in for the last model, which is identical to the "resnet50" model of Fig. 3E; bottom row). In both cases, the networks trained to predict categories on COCO are similar or worse than networks trained to predict categories on ecoset or imagenet, showing that training on COCO is not enough to improve brain alignment compared to other training datasets. The coco-LLM-trained networks outperform all category-trained networks, showing that LLM-training, and not the COCO dataset, explains the good brain alignment of our models. Y-axis shows noise-ceiling corrected RDM correlations averaged across 8 participants; stars denote significant differences; significance threshold set by a 2-tailed t-test across subjects with Benjamini & Hochberg false discovery rate correction; all significant *p*-values are <0.0005).

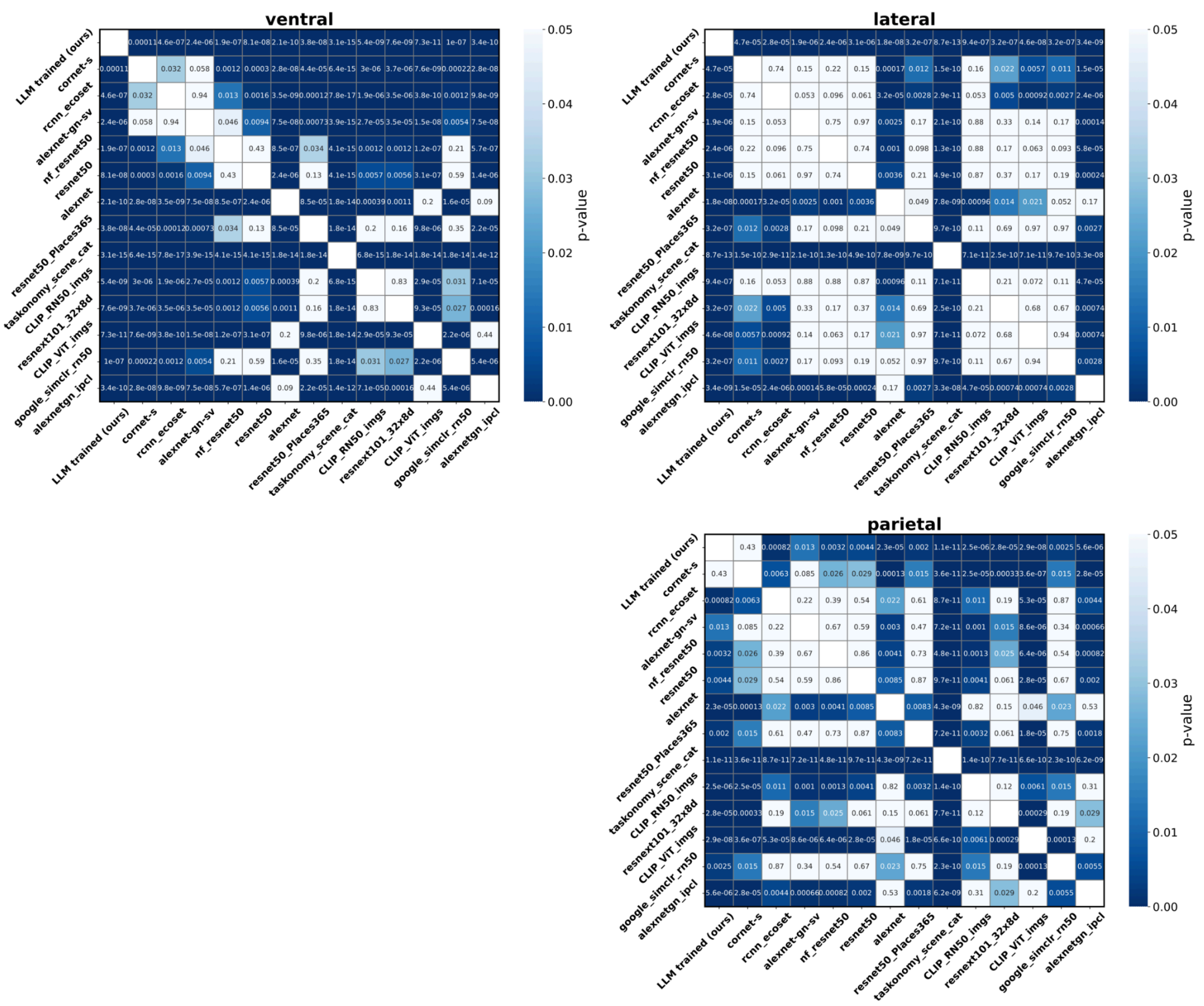

**Supp. Figure 13: All Benjamin & Hochstein false discovery rate corrected p-values for Figure 3E.** 2-tailed *t*-test across the 8 NSD participants. Corrections are conducted across all models, independently for each ROI.